\documentclass{article}
\usepackage{iclr2027_conference,times}
\usepackage{amsmath,amssymb,booktabs,graphicx,xcolor,colortbl,tabularx,float}
\usepackage{hyperref,url}
\usepackage{wrapfig}
\hypersetup{colorlinks=true,linkcolor=teal!65!black,citecolor=teal!65!black,urlcolor=teal!65!black}
\definecolor{tint}{HTML}{E8F4F3}
\definecolor{ink}{HTML}{154452}
\newcommand{\ours}{\textsc{PACT}}

\title{Preserve-and-Compose Training for \\Composed Image Retrieval}
\author{Sehyun Kwon\\
Hanyang University ERICA\\
\texttt{sehyunk@hanyang.ac.kr}}
\iclrfinalcopy
\begin{document}
\raggedbottom
\maketitle
\lhead{Preprint}
\begin{abstract}
Composed image retrieval (CIR) aims to retrieve images that satisfy a user-specified modification while preserving relevant visual content from a reference image. Collecting target images for this purpose is costly, motivating zero-shot CIR methods that instead use target captions as supervision. However, target captions may omit source details that should be preserved. We therefore propose \ours{}, \emph{Preserve-and-Compose Training}, which complements target-caption supervision with visual evidence from the source image. PACT learns from image--text--text (ITT) triplets without target images or gallery updates, aligning composed queries with target captions while preserving source evidence through visual supervision. We further introduce Chord scoring, which combines target similarity with source-relative directional agreement in the frozen image space. Results across four ZS-CIR benchmarks show that combining target-caption supervision with source-image evidence leads to strong retrieval performance across datasets, backbone scales, and external galleries. The code is available on \href{https://github.com/sehyunkwon/PACT}{GitHub}.
\end{abstract}

\section{Introduction}
``Show the same person practicing martial arts'' is not a complete description of an image.
This is the central challenge of composed image retrieval (CIR), which retrieves an image by combining a reference image with a user-specified modification. The instruction specifies an action; the reference image tells the system whom to retrieve. Composed image retrieval (CIR) must combine these complementary inputs: it should change the attributes named in the instruction while retaining relevant visual evidence from the reference. A result that follows the instruction but depicts the wrong person is no more satisfactory than one that preserves the person but ignores the requested action.

Learning this behavior requires supervision about both what should change and what should remain. A matched source--modification--target-image triplet provides a direct example, but finding or generating a target that satisfies both requirements is costly. This motivates learning without explicit target images. Image--caption pairs support visual inversion, and language-based approaches provide inexpensive compositional supervision \citep{saito2023pic2word,gu2024lincir,byun2025rtd}. More recently, source images have been paired with modification instructions and generated target captions (\citealp{li2024mcl}; \citealp{li2026mota}). These alternatives make composition learning possible without collecting a corresponding target image for every instruction.

Target captions nevertheless provide incomplete evidence about the desired result. A caption such as ``a person practicing martial arts'' can express the requested action without specifying the person's appearance or clothing. Those details may still matter for retrieval, even when neither the instruction nor the generated caption mentions them. The source image contains evidence that can complement this textual target. The challenge is therefore not simply to expose a model to an image and a modification, but to use the available supervision to learn a query that reflects both.

To address this gap, we propose \ours{}, \emph{Preserve-and-Compose Training}, which combines caption-endpoint learning with source-image and negative-caption supervision. PACT optimizes a lightweight MLP mapper and LoRA parameters in the text encoder, while keeping the image encoder and gallery embeddings fixed. This design learns composition directly in the vision--language embedding space, without requiring target images or modifying the gallery. It also avoids heavy generative models during retrieval, making the model directly applicable to existing galleries.

\begin{figure}[H]
\centering\includegraphics[width=0.9\linewidth]{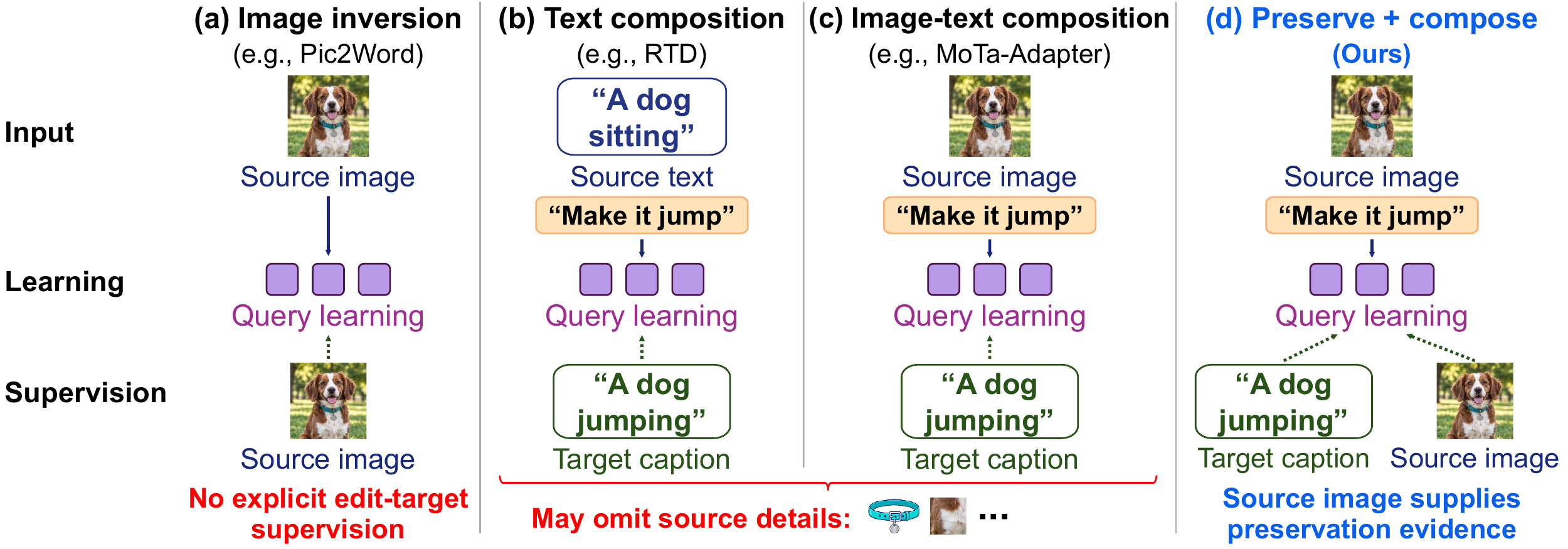}
\caption{\textbf{Supervise both the requested change and the available source evidence.} Training triplets pair a source image with a modification instruction and a target caption, providing a modification target together with a real visual teacher. The diagram characterizes supervision signals, not measured failure rates of all methods in each family.}\label{fig:motivation}
\end{figure}

Figure~\ref{fig:motivation} contrasts the supervision used by different approaches to learn composed queries. Image inversion learns a query representation from the source image without explicit edit-target supervision, while text composition uses a target caption to supervise the desired change. Image--text composition additionally conditions the query on the source image, but still uses the target caption as the primary learning target.

PACT instead assigns complementary roles to the two available sources of supervision. The target caption specifies the desired modification, while the source image provides direct evidence of what should be preserved. Accordingly, each composed query is aligned with its target caption, whereas the shared representation across multiple modifications of the same source is aligned with the source-image embedding. This encourages the query to reflect the requested change without discarding visual information that the target caption leaves unspecified.

We also introduce \emph{Chord scoring} to complement composition learning at retrieval time. Standard cosine similarity measures how well a candidate matches the composed query, but does not account for whether its change from the reference is consistent with the requested modification. Chord augments query--candidate similarity with agreement between the change from the identity query to the modified query and the change from the reference image to each candidate. While related directional matching has been explored previously \citep{tursun2026pdv,wang2025dipnec}, Chord combines this change-based agreement with standard similarity and can be applied to existing query models without retraining. We evaluate both its overall retrieval gains and its ability to resolve ambiguous rankings under cosine similarity.

PACT achieves state-of-the-art or competitive results across four CIR benchmarks, with strong gains on CIRR and CIRCO. Beyond benchmark scores, qualitative retrieval over external image and movie galleries shows that PACT better preserves reference-specific visual identity while following the requested modification, compared with a strong prior method. PACT is also highly efficient to train. The ViT-L/14 model requires less than one hour on a single RTX 3090, while keeping the image encoder and gallery representations frozen. Together, these results demonstrate that PACT combines strong retrieval accuracy, visual preservation, and practical training efficiency.

Our contributions are:
\begin{itemize}
\item We introduce \ours{}, a preserve-and-compose training framework that combines target-caption supervision with source-image grounding, enabling CIR without target-image supervision or gallery updates.
\item We propose \emph{Chord scoring}, a training-free retrieval rule that jointly considers query--candidate similarity and whether the candidate changes from the reference consistently with the requested modification.
\item PACT achieves state-of-the-art or competitive performance across four CIR benchmarks, preserves reference-specific visual identity in large external galleries.
\end{itemize}

\section{Related Work}

\subsection{CIR without target-image supervision}
Supervised CIR typically relies on image--text--image (ITI) triplets, but collecting target images at scale is costly. Zero-shot CIR therefore replaces target-image supervision with more accessible signals. Image--text (IT) methods such as Pic2Word \citep{saito2023pic2word} and SEARLE \citep{baldrati2023zero} learn visual inversion from image--caption pairs, while Context-I2W \citep{tang2024context}, MoA \citep{li2025moa}, HIT \citep{li2025hit}, PrediCIR \citep{tang2025predicir}, and FoCo \citep{zhang2026foco} further improve visual representation or proxy-task design. Language-only approaches include LinCIR \citep{gu2024lincir} and RTD \citep{byun2025rtd}, with RTD explicitly learning from text--text--text (TTT) triplets. Image--text--text (ITT) methods such as MCL \citep{li2024mcl} and MoTa-Adapter \citep{li2026mota} retain the source image while replacing the target image with a caption. CIReVL \citep{karthik2024cirevl} and CoLLM \citep{huynh2025collm} instead employ large language models during retrieval. PACT also learns from ITT triplets, but uses the target caption to supervise the requested composition and the source image as direct visual supervision for what should be preserved.

\subsection{Retrieval scoring for composed queries}
Most CIR methods rank gallery images by cosine similarity to the composed query, while several approaches exploit geometric structure in the embedding space beyond standard cosine matching. Slerp \citep{jang2024slerp} combines image and text representations through spherical interpolation, PDV \citep{tursun2026pdv} uses prompt-induced directions to guide query composition, and DIPNEC \citep{wang2025dipnec} relates modification text to reference-to-candidate visual changes. These approaches incorporate geometric cues into query construction or representation learning. In contrast, we propose \emph{Chord scoring}, a retrieval method for CIR that leaves the learned query representation unchanged and uses reference-relative change only at ranking time, together with standard query--candidate similarity. Chord can therefore be applied to existing CIR models without additional models or retraining, while retaining the same asymptotic retrieval complexity as cosine scoring.
\section{Method}\label{sec:method}

\begin{figure}[t]
\centering
\includegraphics[width=\linewidth]{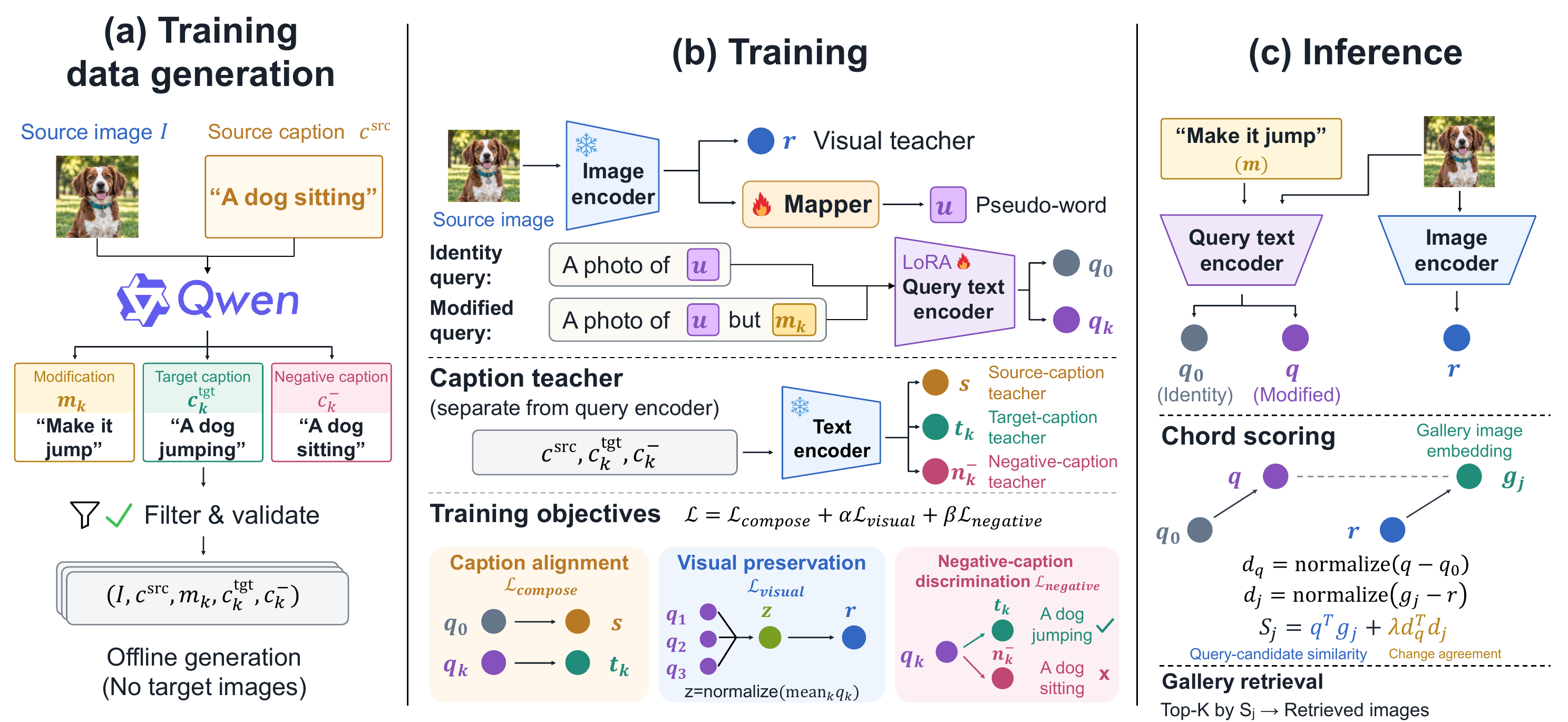}
\caption{\textbf{PACT.}
\textbf{(a)} We construct ITT training data from a source image, modification instructions, target captions, and optional negative captions, without target images.
\textbf{(b)} During training, target captions supervise the composition, source images provide visual supervision for information shared across modifications, and negative captions discourage failed modifications.
\textbf{(c)} At inference, Chord scoring combines query--candidate similarity with reference-relative change for retrieval.}
\label{fig:pipeline}
\end{figure}

PACT learns composed queries from image--text--text (ITT) supervision without requiring target images or modifying gallery representations. Given a source image and a modification instruction, we aim to construct a query that reflects the requested change while retaining visual evidence from the source that may not be expressed in text. Figure~\ref{fig:pipeline} summarizes the overall framework. Section~\ref{sec:data} describes how we construct ITT training data. Section~\ref{sec:compose} introduces the caption-guided composition objective $\mathcal L_{\rm compose}$. Section~\ref{sec:preserve} complements this textual supervision with source-image preservation and negative-caption discrimination through $\mathcal L_{\rm preserve}$. Finally, Section~\ref{sec:chord} presents Chord scoring, our retrieval method for incorporating reference-relative change at inference.

\subsection{Constructing image--text--text training data}\label{sec:data}

Each \emph{image--text--text (ITT) triplet} consists of a source image $I_i$, a modification instruction $m_{i,k}$, and a target caption $c_{i,k}^{\rm tgt}$ describing the desired result. We use CC3M images \citep{sharma2018conceptual} with their associated source captions, including DreamLIP recaptions \citep{zheng2024dreamlip}. This replaces target-image supervision with a textual description of the desired result.

We construct ITT triplets in two ways. First, we create modification--target-caption pairs through rule-based substitution of explicit attributes in source captions. Second, we use Qwen3.5-9B \citep{qwen2026qwen35} to generate additional pairs directly from source images. For each image-conditioned modification, we also request a negative caption $c_{i,k}^{-}$ that preserves relevant source information but does not satisfy the requested modification. We remove malformed or overlength records, duplicate instructions, no-change or invalid generations, and negative captions that fail validation. The resulting corpus contains 313,240 source images and 856,089 modification triplets, with up to three retained modifications per source. Appendix~\ref{app:training} provides further details on generation and filtering.

\subsection{Caption-guided composition: $\mathcal L_{\rm compose}$}
\label{sec:compose}

Consider an ITT triplet $(I,m,c^{\rm tgt})$ with source caption $c^{\rm src}$. We additionally use an identity entry $(I,\varnothing,c^{\rm src})$, allowing the same query model to represent the source before and after modification.

Let $E_I$ and $E_T^0$ denote the frozen CLIP image and text encoders \citep{radford2021clip}, whose outputs lie in a shared $d$-dimensional embedding space. We define
\begin{equation}\label{eq:teachers}
r=E_I(I),\qquad
s=E_T^0(c^{\rm src}),\qquad
t=E_T^0(c^{\rm tgt}).
\end{equation}
Here, $r$, $s$, and $t$ represent the source image, source caption, and target caption, respectively. 

Inspired by Pic2Word \citep{saito2023pic2word}, a lightweight MLP mapper $M_\phi$ converts the source-image feature into an image-dependent pseudo-word embedding $u=M_\phi(E_I(I))$. We define $\mathcal P(u,m)$ as the prompt embedding sequence corresponding to a template such as ``a photo of $[u]$ but $[m]$'', where $[u]$ is replaced by $u$ and $[m]$ contains the token embeddings of the modification instruction. The resulting query representation is
\begin{equation}\label{eq:queries}
q
=
f_{\theta,\phi}(I,m)
=
E_T^\theta\!\left(\mathcal P(u,m)\right),
\end{equation}
where $E_T^\theta$ is the CLIP text encoder adapted with attention LoRA \citep{hu2022lora}. For the identity query $q_0=f_{\theta,\phi}(I,\varnothing)$, the modification text is omitted. The pseudo-word carries information from the source image, while the modification text specifies how that content should change. The image encoder and teacher text encoder remain frozen; only the mapper and LoRA parameters are learned.

We align identity queries $q_0$ with source-caption embeddings $s$, and modified queries $q$ with target-caption embeddings $t$. For a minibatch of $B$ such pairs, we stack the query and caption embeddings into $Q,T\in\mathbb R^{B\times d}$, respectively, with matching rows as positives and all other in-batch pairs as negatives. We minimize
\begin{equation}\label{eq:compose}
\mathcal L_{\rm compose}
=
\mathcal C_{\tau_c}(Q,T),
\end{equation}
where
\begin{equation}\label{eq:symmetric_ce}
\mathcal C_{\tau_{c}}(Q,T)
=
-\frac{1}{2B}
\sum_{a=1}^{B}
\left[
\log
\frac{\exp(Q_a^\top T_a/\tau_c)}
{\sum_{b=1}^{B}\exp(Q_a^\top T_b/\tau_c)}
+
\log
\frac{\exp(T_a^\top Q_a/\tau_c)}
{\sum_{b=1}^{B}\exp(T_a^\top Q_b/\tau_c)}
\right].
\end{equation}
This objective learns the source state and desired composition from captions. However, captions may omit source details, motivating the preservation objective in Section~\ref{sec:preserve}.

\newpage
\subsection{Preservation objective: $\mathcal L_{\rm preserve}$}
\label{sec:preserve}

For a source image $I_i$ with retained modifications
$\{m_{i,k}\}_{k=1}^{K_i}$, we denote the corresponding queries by
$q_{i,k}=f_{\theta,\phi}(I_i,m_{i,k})$ and the source-image representation by
$r_i=E_I(I_i)$.

\paragraph{Visual preservation.}
Target captions may omit visual details that should remain after modification.
Rather than aligning each modified query directly with the source image, which
could interfere with the requested change, we supervise the information shared
across multiple modifications of the same source.

For $\mathcal L_{\rm visual}$, we form a preservation batch in which some
source images contribute multiple modified queries. Let $\mathcal K_i$ denote
the modifications of source $I_i$ appearing in the batch. For
$|\mathcal K_i|\ge2$, we define their shared representation as
\begin{equation}\label{eq:pooled}
z_i
=
\operatorname{normalize}
\left(
\frac{1}{|\mathcal K_i|}
\sum_{k\in\mathcal K_i} q_{i,k}
\right).
\end{equation}
The identity query is excluded, and sources appearing only once do not
contribute to this loss.

Let $r_i=E_I(I_i)$ denote the frozen visual teacher for source $I_i$, and let
$\mathcal S$ be the source images represented in the preservation batch.
We contrastively align $z_i$ with its source-image teacher:
\begin{equation}\label{eq:visual}
\mathcal L_{\rm visual}
=
-\operatorname{mean}_{i:\,|\mathcal K_i|\ge2}
\log
\frac{
\exp(z_i^\top r_i/\tau_v)
}{
\sum_{j\in\mathcal S}\exp(z_i^\top r_j/\tau_v)
}.
\end{equation}
Thus, the source image supervises visual evidence shared across different
modifications, including details that may be absent from any individual target
caption. For clarity, Equation~\ref{eq:visual} shows one visual teacher per
source; the implementation retains repeated and near-identical positives in
the preservation batch, as detailed in Appendix~\ref{app:objectives}.

\paragraph{Negative-caption discrimination.}
Preserving source evidence should not prevent the requested modification. For
a modification $m_{i,k}$ with target caption $c_{i,k}^{\rm tgt}$ and valid
negative caption $c_{i,k}^{-}$, define
\begin{equation}
t_{i,k}=E_T^0(c_{i,k}^{\rm tgt}),
\qquad
n_{i,k}^{-}=E_T^0(c_{i,k}^{-}).
\end{equation}
The negative caption preserves relevant source information but fails to satisfy the requested modification. Let $\mathcal V$ denote the set of entries with negative captions; we minimize
\begin{equation}\label{eq:negative}
\mathcal L_{\rm negative}
=
-\frac{1}{|\mathcal V|}
\sum_{(i,k)\in\mathcal V}
\log
\frac{
\exp(q_{i,k}^\top t_{i,k}/\tau_n)
}{
\exp(q_{i,k}^\top t_{i,k}/\tau_n)
+
\exp(q_{i,k}^\top n_{i,k}^{-}/\tau_n)
}.
\end{equation}

The two terms play complementary roles: $\mathcal L_{\rm visual}$ preserves
source evidence, while $\mathcal L_{\rm negative}$ discourages failure to make
the requested change. We define
\begin{equation}\label{eq:preserve}
\mathcal L_{\rm preserve}
=
\alpha\mathcal L_{\rm visual}
+
\beta\mathcal L_{\rm negative},
\end{equation}
and optimize
\begin{equation}\label{eq:full}
\boxed{
\mathcal L_{\rm PACT}
=
\mathcal L_{\rm compose}
+
\mathcal L_{\rm preserve}
}.
\end{equation}
No explicit transition-alignment loss is required.

\subsection{Inference with Chord scoring}\label{sec:chord}

Standard CIR retrieval ranks gallery images by cosine similarity to the composed query. While this measures how well a candidate matches the desired result, it does not consider whether the candidate differs from the reference consistently with the requested modification. We therefore introduce \emph{Chord scoring}, which incorporates this reference-relative information at ranking time.

Given a reference image $I$, modification $m$, and gallery image $G_j$, we compute
\begin{equation}
r=E_I(I),\qquad
q=f_{\theta,\phi}(I,m),\qquad
q_0=f_{\theta,\phi}(I,\varnothing),\qquad
g_j=E_I(G_j),
\end{equation}
where $f_{\theta,\phi}$ denotes the query encoder formed by $M_\phi$ and $E_T^\theta$.
Chord scoring combines standard query--candidate similarity with agreement between two changes in the shared embedding space:
\begin{equation}\label{eq:main_score}
S_j
=
\cos(q,g_j)
+
\lambda\,
\cos(q-q_0,\;g_j-r).
\end{equation}
The first term measures similarity between the composed query and candidate, while the second measures whether the candidate changes from the reference consistently with the query change. Because these differences connect pairs of unit embeddings, they form chords of the embedding sphere, motivating the name \emph{Chord scoring}. The cosine term compares the directions of these two changes, independent of their magnitudes.

The weight $\lambda\ge0$ controls the contribution of this reference-relative term and is selected on validation data for each query model. Chord leaves both the learned query representation and gallery embeddings unchanged, allowing it to be applied without an additional model or retraining. For a gallery of $N$ $d$-dimensional embeddings, Chord retains the same \(O(Nd)\) asymptotic complexity as cosine scoring, with only an additional candidate-wise cosine computation.

\section{Experiments}\label{sec:experiments}

\subsection{Evaluation protocol}

We evaluate PACT on CIRR \citep{liu2021cirr}, CIRCO \citep{baldrati2023zero},
FashionIQ \citep{wu2019fashioniq}, and GeneCIS \citep{vaze2023genecis}.
Unless otherwise stated, all comparisons use ViT-L/14, and PACT results are
averaged over five independent runs with different random seeds.
We report official test results on CIRR and CIRCO and validation results on
FashionIQ and GeneCIS, with FashionIQ recall averaged equally across categories.
Baselines follow the retrieval procedures of their source evaluations, whereas
PACT uses Chord scoring; accordingly, the main benchmark tables compare complete
retrieval systems rather than training objectives in isolation.
For any model evaluated with Chord scoring, we select a single weight $\lambda$
on validation using the mean of CIRR R@1, CIRCO mAP@5, and FashionIQ
category-average R@10, and fix it for test and transfer evaluation.
Section~\ref{sec:chord} separately isolates the effect of Chord scoring on
models for which checkpoints are available.
Appendix~\ref{app:evaluation} provides full evaluation details.

\subsection{Benchmark performance}\label{sec:benchmarks}

\begingroup
\setlength{\intextsep}{5pt}
\renewcommand{\arraystretch}{0.92}
% Register the papers linked by the venue labels.
\nocite{
baldrati2023zero,
byun2025rtd,
gu2024lincir,
karthik2024cirevl,
li2025hit,
li2025moa,
li2026mota,
saito2023pic2word,
tang2024context,
tang2025predicir,
tang2026deminds,
he2026flowcir}

\begin{table}[H]
\centering
\scriptsize
\setlength{\tabcolsep}{3pt}
\caption{\textbf{CIRR/CIRCO test and FashionIQ validation, ViT-L/14.}
Published baselines and official PACT test results; FashionIQ reports
category-average recall.}
\label{tab:requested1}\label{tab:requested2}
\vspace{0.1cm}
\resizebox{\linewidth}{!}{
\begin{tabular}{lrrrrrrrr}
\toprule
Method &
\multicolumn{3}{c}{CIRR test} &
\multicolumn{3}{c}{CIRCO test} &
\multicolumn{2}{c}{FashionIQ val}\\
\cmidrule(lr){2-4}
\cmidrule(lr){5-7}
\cmidrule(lr){8-9}
& R@1 & R@5 & R@10
& mAP@5 & mAP@10 & mAP@25
& Avg R@10 & Avg R@50\\
\midrule

Pic2Word \hyperlink{cite.saito2023pic2word}{(CVPR 2023)}
&23.90&51.70&65.30&8.72&9.51&10.64&24.70&43.70\\

SEARLE-XL \hyperlink{cite.baldrati2023zero}{(ICCV 2023)}
&24.24&52.48&66.29&11.68&12.73&14.33&25.56&46.23\\

Context-I2W \hyperlink{cite.tang2024context}{(AAAI 2024)}
&25.60&55.10&68.50&13.00&14.60&16.10&27.80&48.90\\

CIReVL (GPT-3.5) \hyperlink{cite.karthik2024cirevl}{(ICLR 2024)}
&24.55&52.31&64.92&18.57&19.01&20.89&28.55&48.57\\

LinCIR \hyperlink{cite.gu2024lincir}{(CVPR 2024)}
&25.04&53.25&66.68&12.59&13.58&15.00&26.28&46.49\\

PrediCIR \hyperlink{cite.tang2025predicir}{(CVPR 2025)}
&27.20&57.00&70.20&15.70&17.10&18.60&30.10&52.30\\

MoA \hyperlink{cite.li2025moa}{(SIGIR 2025)}
&27.10&56.50&69.20&15.30&17.10&18.50&30.10&51.30\\

HIT$^\ast$ \hyperlink{cite.li2025hit}{(ICCV 2025)}
&27.90&57.60&70.50&15.50&16.70&18.90&30.30&51.00\\

RTD \hyperlink{cite.byun2025rtd}{(ICCV 2025)}
&26.63&56.17&68.96&17.11&18.11&20.06&30.24&51.08\\

MoTa-Adapter \hyperlink{cite.li2026mota}{(AAAI 2026)}
&28.02&58.62&71.06&18.22&19.46&--&28.76&49.53\\

De-MINDS \hyperlink{cite.tang2026deminds}{(AAAI 2026)}
&30.00&59.70&\textbf{74.50}
&21.00&22.40&22.70
&32.20&\textbf{54.50}\\

FlowCIR \hyperlink{cite.he2026flowcir}{(ECCV 2026)}
&26.20&56.10&68.60
&14.90&15.70&17.30
&29.70&48.90\\

\rowcolor{tint}
PACT (ours)
&\textbf{31.24}&\textbf{60.39}&71.66
&\textbf{23.59}&\textbf{24.52}&\textbf{26.89}
&\textbf{32.43}&52.29\\

\bottomrule
\end{tabular}
}
\end{table}

\endgroup

\paragraph{Composed image retrieval.}
PACT achieves strong performance across all three benchmarks in
Table~\ref{tab:requested1}.
The largest gains are observed on CIRCO, where PACT achieves the best
results among the compared methods across all reported metrics.
At mAP@5, PACT reaches 23.59, exceeding the strongest baseline, De-MINDS,
by 2.59 points and CIReVL by 5.02 points; at mAP@25, the margin over
De-MINDS increases to 4.19 points.
On CIRR, PACT achieves the highest R@1 (31.24) and R@5 (60.39), while
De-MINDS remains higher at R@10.
On FashionIQ, PACT achieves the highest category-average R@10 (32.43),
while De-MINDS remains higher at R@50 (54.50 versus 52.29).

\begingroup
\setlength{\intextsep}{5pt}
\renewcommand{\arraystretch}{0.92}
% Register the papers linked by the venue labels.
\nocite{baldrati2023zero,
byun2025rtd,
gu2024lincir,
karthik2024cirevl,
li2026mota,
saito2023pic2word,
zhang2026diffcomp}

\begin{table}[H]
\centering
\scriptsize
\setlength{\tabcolsep}{3pt}
\caption{\textbf{GeneCIS validation, ViT-L/14.}
Local and published baseline results; -- denotes unreported metrics.}
\label{tab:requested3}
\vspace{0.1cm}
\resizebox{\linewidth}{!}{
\begin{tabular}{lrrrrrrrrrrrr}
\toprule
Method
& \multicolumn{3}{c}{Focus}
& \multicolumn{3}{c}{Change}
& \multicolumn{3}{c}{Attribute}
& \multicolumn{3}{c}{Object}\\
\cmidrule(lr){2-4}
\cmidrule(lr){5-7}
\cmidrule(lr){8-10}
\cmidrule(lr){11-13}
& R@1 & R@2 & R@3
& R@1 & R@2 & R@3
& R@1 & R@2 & R@3
& R@1 & R@2 & R@3\\
\midrule

Pic2Word \hyperlink{cite.saito2023pic2word}{(CVPR 2023)}
&13.36&24.84&34.66
&13.95&24.67&34.51
&14.93&26.27&35.87
&12.37&23.24&33.29\\

SEARLE-XL \hyperlink{cite.baldrati2023zero}{(ICCV 2023)}
&14.29&25.39&36.09
&14.36&25.00&33.67
&16.81&27.56&37.59
&11.84&22.83&32.17\\

CIReVL (GPT-3.5) \hyperlink{cite.karthik2024cirevl}{(ICLR 2024)}
&15.90&26.80&36.25
&15.80&27.45&36.40
&16.95&28.90&38.60
&14.75&25.35&34.05\\

LinCIR \hyperlink{cite.gu2024lincir}{(CVPR 2024)}
&14.55&26.18&35.71
&15.08&26.77&36.53
&16.37&29.17&38.61
&13.27&23.78&33.62\\

RTD \hyperlink{cite.byun2025rtd}{(ICCV 2025)}
&15.38&26.17&36.61
&15.86&27.38&36.74
&17.47&29.42&39.91
&13.78&24.13&33.44\\

MoTa-Adapter \hyperlink{cite.li2026mota}{(AAAI 2026)}
&\multicolumn{12}{c}{
Reported GeneCIS R@1 summary: 16.7; task marginals unreported
}\\

DiffComp \hyperlink{cite.zhang2026diffcomp}{(CVPR 2026)}
&17.95&\textbf{29.73}&--
&15.95&26.68&--
&19.00&30.39&--
&14.89&\textbf{26.02}&--\\

\rowcolor{tint}
PACT (ours)
&\textbf{18.15}&28.54&\textbf{38.96}
&\textbf{16.96}&\textbf{28.73}&\textbf{38.77}
&\textbf{20.06}&\textbf{31.90}&\textbf{42.18}
&\textbf{15.06}&25.38&\textbf{35.55}\\

\bottomrule
\end{tabular}
}
\end{table}

\endgroup

\paragraph{Transfer to conditional image similarity.}
GeneCIS evaluates conditional similarity under conditions that focus on or
change objects and attributes.
PACT achieves the highest R@1 across all four reported marginals in
Table~\ref{tab:requested3}.
PACT consistently outperforms the locally evaluated Pic2Word, LinCIR,
and RTD models across the displayed recall metrics.
Compared with DiffComp, PACT is higher on all four R@1 metrics, while
DiffComp is higher on Focus R@2 (29.73 versus 28.54) and Object R@2
(26.02 versus 25.38).
The Chord weight is transferred from CIRR, CIRCO, and FashionIQ without
tuning on GeneCIS, indicating that the learned retrieval behavior generalizes
beyond the standard CIR benchmarks.

\subsection{Ablation of training objectives and retrieval scoring}
\label{sec:ablations}

\begin{table}[H]
\centering
\small
\setlength{\tabcolsep}{6pt}
\caption{\textbf{Ablation of PACT training objectives and retrieval scoring.}
Validation performance with ViT-L/14.
The mean averages CIRR R@1, CIRCO mAP@5, and FashionIQ R@10.
The cosine comparison uses the same checkpoints as the corresponding
Chord results.}
\label{tab:ablation}
\vspace{0.1cm}
\begin{tabular}{lrrrr}
\toprule
Training configuration & CIRR & CIRCO & FIQ & Mean\\
& R@1 & mAP@5 & R@10 & \\
\midrule

\multicolumn{5}{l}{\textit{Chord scoring}}\\
\rowcolor{tint}
\textbf{PACT (full)} &
\textbf{32.50} &
\textbf{24.18} &
\textbf{32.43} &
\textbf{29.70}\\
$\mathcal L_{\rm compose}$ only &
29.47 & 23.80 & 29.19 & 27.49\\
$\mathcal L_{\rm preserve}$ only &
28.58 & 16.12 & 29.45 & 24.72\\
$\mathcal L_{\rm compose}+\alpha\mathcal L_{\rm visual}$ &
30.40 & 23.64 & 30.47 & 28.17\\
$\mathcal L_{\rm compose}+\beta\mathcal L_{\rm negative}$ &
31.26 & 22.83 & 29.23 & 27.77\\

\midrule
\multicolumn{5}{l}{\textit{Cosine scoring (matched checkpoints)}}\\
\rowcolor{tint}
\textbf{PACT (full)} &
\textbf{25.76} &
\textbf{18.68} &
\textbf{28.29} &
\textbf{24.24}\\
$\mathcal L_{\rm compose}$ only &
23.10 & 16.92 & 24.55 & 21.52\\

\bottomrule
\end{tabular}
\end{table}

\paragraph{Composition and preservation provide complementary supervision.}
With Chord scoring, $\mathcal L_{\rm compose}$ alone achieves a
three-benchmark mean of 27.49, outperforming
$\mathcal L_{\rm preserve}$ alone (24.72), showing that preservation cannot
replace supervision for the requested change.
Adding either $\mathcal L_{\rm visual}$ or $\mathcal L_{\rm negative}$ gives
a partial gain, while combining both yields the best mean of 29.70 and improves
all three benchmarks over composition-only training.
Neither auxiliary term alone recovers the full improvement, supporting their
complementary roles in preserving source evidence and enforcing the requested
modification.
Thus, composition provides the primary learning signal, with preservation
providing additional supervision.

\paragraph{PACT gains persist without Chord.}
To separate training from retrieval, we evaluate the same checkpoints with
standard cosine similarity.
PACT improves the mean over composition-only training by 2.72 points with
cosine (21.52$\rightarrow$24.24) and by 2.21 points with Chord
(27.49$\rightarrow$29.70).
The gains under both retrieval rules show that the benefit of PACT
is not specific to Chord.
The improvement is also consistent across CIRR, CIRCO, and FashionIQ under
both scoring rules.
Chord adds 5.46 points to PACT and 5.97 to composition-only training.
Thus, PACT improves the learned query representation, while Chord provides an
additional inference-time gain.
Per-dataset scoring comparisons are provided in
Appendix~\ref{app:chordanalysis}.

\subsection{Effectiveness and behavior of Chord scoring}
\label{sec:chordanalysis}

\begin{figure}[H]
\centering
\includegraphics[width=0.85\linewidth]{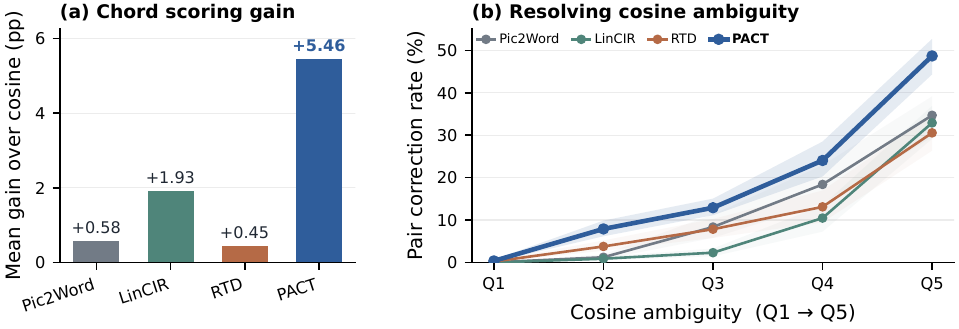}
\caption{\textbf{Chord scoring improves retrieval and resolves ambiguous
cosine rankings.}
\textbf{(a)} Mean improvement over cosine across CIRR R@1, CIRCO mAP@5,
and FashionIQ R@10, using one validation-selected $\lambda$ per model.
\textbf{(b)} Correction rate for positive--candidate pairs misordered by
cosine similarity at a common $\lambda=0.34$.
Q1--Q5 correspond to decreasing cosine deficits; shading denotes 95\%
reference-cluster bootstrap intervals.
Appendix~\ref{app:chordanalysis} provides full details.}
\label{fig:intrinsic}
\end{figure}

\paragraph{Chord scoring generalizes across query models.}
Figure~\ref{fig:intrinsic}a applies Chord post hoc to Pic2Word, LinCIR, RTD,
and PACT without retraining.
Relative to cosine retrieval, it improves the three-benchmark mean by
0.58, 1.93, 0.45, and 5.46 points, respectively.
Thus, reference-relative scoring benefits multiple query models, with the
largest gain observed for PACT.

\paragraph{Reference-relative change resolves cosine ambiguity.}
For a positive $p$ and the highest-cosine non-positive candidate $n$, let
\[
\delta=\cos(q,g_n)-\cos(q,g_p)>0,\qquad
D_j=\cos(q-q_0,g_j-r).
\]
Then
\begin{equation}
S_p-S_n=-\delta+\lambda(D_p-D_n).
\end{equation}
Chord reverses the cosine ordering when the change-agreement term outweighs
the cosine deficit.
Figure~\ref{fig:intrinsic}b shows higher correction rates for smaller
deficits, indicating that reference-relative change is most useful when
cosine rankings are ambiguous.
This trend is consistent across all four query models, suggesting that the
directional term acts as a complementary signal rather than replacing
query--candidate similarity.
In particular, it can break near-ties by favoring candidates whose change
from the reference better agrees with the requested query change.

\subsection{Model size and efficiency}\label{sec:cost}

\begin{wrapfigure}{r}{0.5\textwidth}
    \vspace{-0.7\baselineskip}
    \centering
    \includegraphics[width=\linewidth]{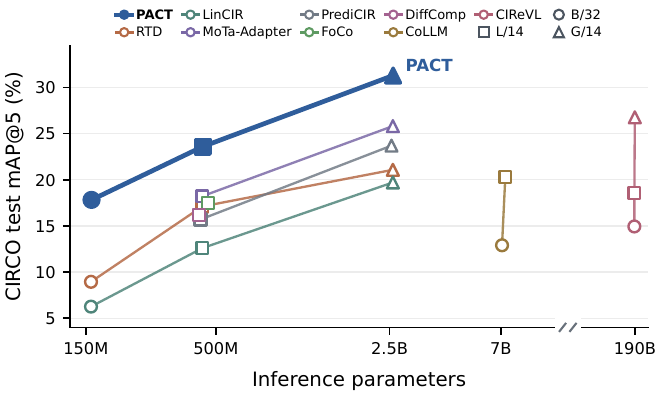}
    \caption{\textbf{Performance versus model size.} PACT remains strong with a compact inference model.}
    \label{fig:params}
    \vspace{-0.4\baselineskip}
\end{wrapfigure}

\paragraph{Strong retrieval with a compact inference model.}
PACT-L/14 achieves 23.59 CIRCO test mAP@5 with approximately 443M inference
parameters.
It outperforms CoLLM-L/14 (20.30), CIReVL-L/14 (18.57),
LinCIR-G/14 (19.71), and RTD-G/14 (21.08), while nearly matching
PrediCIR-G/14 (23.70) with at least $5.7\times$ fewer parameters.
Notably, PACT-L/14 remains competitive with methods using substantially larger
backbones or additional language-model components, despite retaining a compact
CLIP-based inference pipeline.
PACT also outperforms MCL variants using 2.975--7.068B parameters, which
achieve 14.55--17.67 mAP@5 \citep{li2024mcl}.
Across these comparisons, increasing inference model size does not by itself
translate into stronger retrieval performance.
These results show that PACT achieves strong retrieval without requiring a
large language model at inference.

\paragraph{Training efficiency.}
Using a batch size of 512, PACT with a ViT-L/14 backbone completes a 333-step
identity warmup followed by 1,000 joint-training steps in only
\textbf{\(\sim\)52 min} on a single \textbf{RTX 3090}, with
\textbf{8.19 GiB} peak allocated GPU memory.
This efficiency follows from keeping the image encoder frozen and adapting
only lightweight query-side components, without target-image supervision or
gallery re-encoding.
Thus, PACT can substantially adapt a large CLIP retrieval model with modest
training compute and memory.
The measurement excludes evaluation, checkpoint saving, and offline data
preparation.

\begin{figure}[t]
\centering
\includegraphics[width=\linewidth]{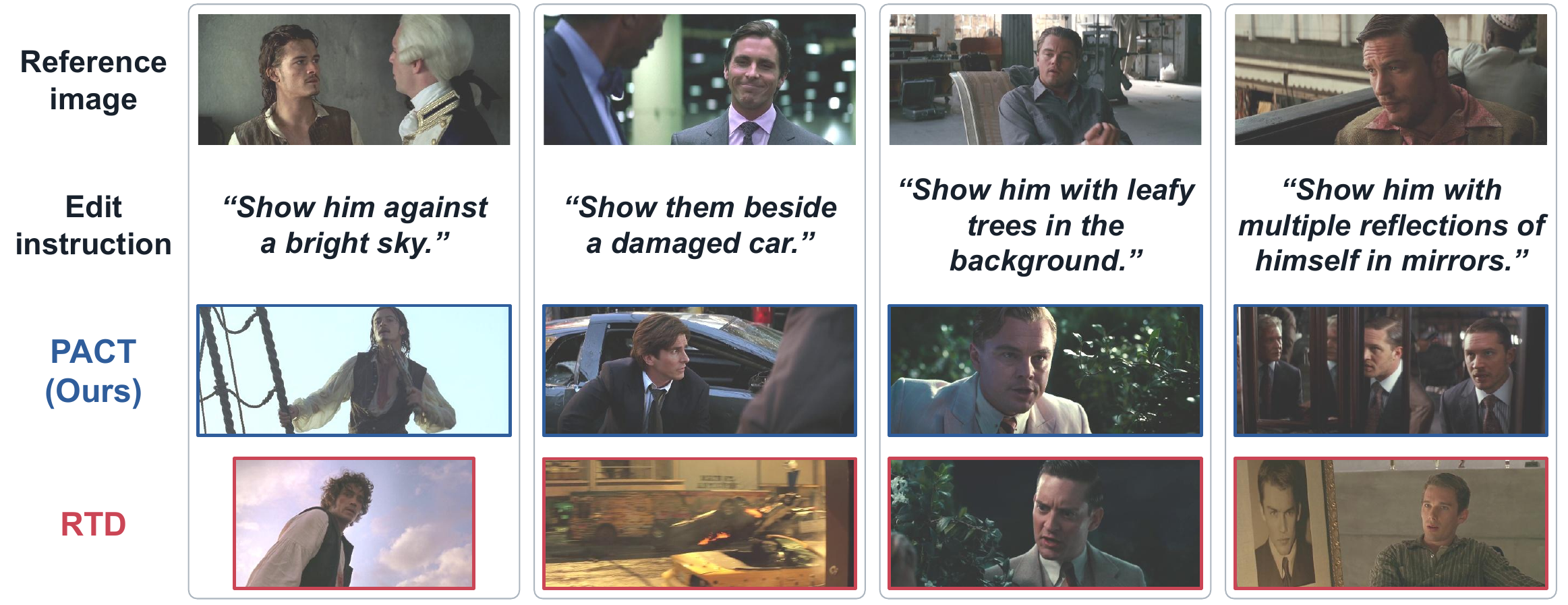}
\caption{\textbf{Reference-dependent retrieval on MovieNet.}
Selected top-1 disagreements between PACT and RTD.
PACT better preserves the referenced identity while following scene- and
background-level modifications.}
\label{fig:qual}
\end{figure}

\subsection{Reference preservation beyond the benchmarks}
\label{sec:qualitative}

\noindent
\begin{minipage}[t]{0.58\linewidth}
\vspace{0pt}
Standard CIR benchmarks do not directly measure whether reference-specific
visual identity is preserved after an edit.
We therefore construct a MovieNet~\citep{huang2020movienet} evaluation set where edits
change the scene, background, or action while the referenced person remains
unchanged.
We measure whether the Top-1 retrieval contains the same annotated identity
as the reference; evaluation details are provided in Appendix~\ref{app:movienet}.
\end{minipage}
\hfill
\begin{minipage}[t]{0.40\linewidth}
\vspace{0pt}
\centering
\small

\refstepcounter{table}
\label{tab:movienet_preserve}
\textbf{Table~\thetable: MovieNet Top-1 identity (\%).}\\[3pt]

\setlength{\tabcolsep}{4pt}
\begin{tabular}{lcc}
\toprule
Method & Cosine & Chord \\
\midrule
RTD & 64.0 & 64.5 \\
\midrule
$\mathcal L_{\rm compose}$ & 58.0 & 62.0 \\
$+\alpha\mathcal L_{\rm visual}$ & 78.5 & 81.5 \\
\rowcolor{tint}
\textbf{PACT (full)} & \textbf{80.5} & \textbf{82.5} \\
\bottomrule
\end{tabular}
\end{minipage}

\vspace{0.3em}

Adding $\mathcal L_{\rm visual}$ to composition substantially improves
identity preservation under both cosine and Chord retrieval.
Full PACT retains this gain and also exceeds RTD, supporting the intended role
of $\mathcal L_{\rm preserve}$ in preserving reference-specific visual evidence.
Figure~\ref{fig:qual} shows complementary qualitative examples from a separate
MovieNet gallery (Appendix~\ref{app:qualitative}).
PACT more consistently retains the reference identity while following the
edit, illustrating the preservation behavior measured above.

\section{Conclusion}

% \paragraph{Limitations.}
% PACT relies on automatically constructed target and negative captions, whose
% semantic quality is imperfect.
% The visual objective encourages source preservation but does not explicitly
% identify which attributes should remain unchanged.
% Our ablations use a single seed, and the external-gallery examples are curated
% rather than intended as an unbiased success-rate estimate.

% \paragraph{Conclusion.}
We introduced PACT, a preserve-and-compose training framework for composed
image retrieval without target-image supervision or gallery updates.
PACT aligns composed queries with target captions while using the source image
to preserve visual evidence that may be omitted from text, with negative-caption
discrimination further encouraging the requested modification.
At inference, Chord scoring complements query--candidate similarity with
reference-relative change agreement and improves multiple query models without
retraining.
PACT achieves strong performance across CIRR, CIRCO, FashionIQ, and GeneCIS,
with particularly strong results on CIRCO.
Qualitative retrieval on external movie galleries further illustrates that PACT
can follow scene-level edits while retaining reference-specific visual identity.
Overall, the results support jointly modeling what should change and what should
be preserved for effective composed image retrieval.
\clearpage

\subsection*{AI use statement}
We used Qwen models to generate training instructions and captions, as detailed in Appendix~\ref{app:training}. We also used Qwen3.5-27B to generate and validate MovieNet evaluation queries and assess modification fulfillment with model identities hidden. Identity preservation was evaluated using official MovieNet annotations. Additionally, we used generative AI tools to brainstorm research ideas and assist with experimental design.

\subsection*{Ethics statement}
Our experiments use existing datasets and pretrained models, whose
biases and errors may affect generated supervision and retrieval
results. Automated filtering and evaluation do not eliminate these
limitations. Person-related retrieval also raises privacy concerns
if applied to tracking individuals without consent. Our MovieNet
experiments evaluate character preservation within movie frames
and do not establish suitability for real-world person identification.

\subsection*{Reproducibility statement}
Section~\ref{sec:method} describes the model, training objectives,
and Chord scoring. Appendices~\ref{app:training}
and~\ref{app:implementation} detail data construction, filtering,
and optimization. The appendix also documents evaluation protocols,
baseline sources, random seeds, scoring-parameter selection,
variation across training runs, and MovieNet evaluation procedures.
\bibliography{references}
\bibliographystyle{iclr2027_conference}
\clearpage
\appendix

\section{Training data construction}\label{app:training}

\subsection{Source images and supervision}
We construct image--text--text triplets from CC3M images
\citep{sharma2018conceptual} and their associated source captions, including
DreamLIP recaptions \citep{zheng2024dreamlip}. Each triplet contains a source
image, a modification instruction, and a target caption. The source image
provides visual supervision, while the target caption describes the requested
composition. Target images are not required.

The training corpus contains 313,240 source images and 856,089 modification
triplets. We additionally form one identity entry per source, pairing the
image with its source caption and an empty modification. The resulting
composition-training pool contains 1,169,329 entries. Identity targets use
the source captions associated with the images. Data generation and feature
preparation are performed offline.

\subsection{Modification generation}
Caption-based construction combines lexical substitutions of explicit
attributes, including color, material, pattern, and count, with language-model
generation. Qwen3-8B \citep{qwen2025qwen3} phrases exact-span modifications
as natural instructions, while Qwen3.5-27B \citep{qwen2026qwen35} proposes
instruction--target-caption pairs from source captions. Together, these
routes provide one caption-based modification record per source.
Image-conditioned generation uses Qwen3.5-9B to produce further
modification--target-caption pairs directly from source images, along with
negative captions. This route contributes 542,849 eligible modification
records, giving 856,089 triplets in total.

\subsection{Filtering and negative captions}
Caption-based modifications undergo consistency and token-length checks.
Image-conditioned proposals contain a source description, a modification,
and a target caption. We remove malformed outputs, duplicate instructions,
no-change modifications, overlength text, and records explicitly rejected
by checks of source support, modification applicability, or target-caption
consistency. The generated source description is used for these checks;
it does not replace the source caption used as the identity target.

The negative-caption objective uses captions that preserve source information
but fail to satisfy the modification. A negative is eligible only when its
validity flag is true, it is nonempty, it differs from the target caption,
and it fits the text-token limit. Negatives that instead violate the intended
preservation constraint are excluded. These automatic filters establish
eligibility for training; they do not constitute human verification of every
retained triplet.

\section{Implementation details}\label{app:implementation}

\subsection{Backbones and optimization}
PACT uses pretrained vision--language encoders with ViT backbones
\citep{dosovitskiy2021vit}. ViT-B/32 and ViT-L/14 initialize from OpenAI CLIP
\citep{radford2021clip}. ViT-G/14 denotes OpenCLIP ViT-bigG/14
\citep{cherti2023openclip}, pretrained on the LAION-2B English subset
\citep{schuhmann2022laion}; the pretrained checkpoint is
\texttt{CLIP-ViT-bigG-14-laion2B-39B-b160k}.
The image encoder and teacher text encoder remain frozen. Learning adapts
the query-side mapper and text-encoder attention
LoRA parameters.

Training comprises 333 identity-warmup updates followed by 1,000 joint
updates. During identity warmup, the text encoder is frozen and identity
queries are aligned with source captions. The warmup uses symmetric
contrastive cross-entropy with temperature 0.07 and masks off-diagonal
source-caption similarities above 0.999. Joint training uses the composition
objective in Equation~\ref{eq:compose} together with the preservation
objective in Equation~\ref{eq:preserve}.

We use AdamW \citep{loshchilov2019adamw} with FP32 optimizer moments and BF16
trainable weights. Learning rates increase linearly over the first 100
updates of each stage and remain constant thereafter. Gradient replay
\citep{gao2021gradcache} processes microbatches of 32 while retaining the
full contrastive batch. Table~\ref{tab:recipe} summarizes the ViT-L/14
configuration.

\begin{table}[htbp]
\centering\small
\caption{\textbf{ViT-L/14 training configuration.}}
\label{tab:recipe}
\begin{tabular}{lr}
\toprule
Setting & Value\\
\midrule
Identity / joint updates & 333 / 1,000\\
Composition / preservation batch size & 512 / 512\\
Replay microbatch size & 32\\
Warmup mapper learning rate & $10^{-4}$\\
Joint mapper / text LoRA learning rate & $3\times10^{-5}$ / $10^{-5}$\\
Weight decay & 0.001\\
Text attention LoRA rank & 16\\
Composition temperature $\tau_c$ & 0.05\\
Visual / negative temperature $\tau_v,\tau_n$ & 0.03 / 0.07\\
Preservation weights $\alpha,\beta$ & 1 / 1\\
Trainable weights / optimizer moments & BF16 / FP32\\
\bottomrule
\end{tabular}
\end{table}

\subsection{Composition and preservation sampling}\label{app:objectives}
Each composition batch contains 512 entries drawn independently and uniformly
with replacement from the union of identity and modification records.
Identity frequency follows the composition of this pool; the sampler imposes
neither a fixed identity quota nor identity--modification pairing within a
batch. In joint training, Equation~\ref{eq:symmetric_ce} treats matched
query--caption positions as positives and all off-diagonal positions as
negatives, without an additional same-source or duplicate-caption mask.

An independently sampled preservation batch contains 256 caption-based and
256 image-conditioned modification records. Image-conditioned records are
sampled with replacement. Of the caption-based records, 128 are selected
to share sources with image-conditioned records, and 128 are sampled
independently. This construction provides groups of modified queries for
the visual objective. The composition and preservation losses are combined
before the optimizer update. Because sampling is with replacement, the
fixed update schedule does not correspond to an epoch over the eligible
training pool.

\subsection{Visual supervision with repeated positives}
Equation~\ref{eq:visual} presents the visual objective with one teacher per
source. Here we specify its implementation when the preservation batch
contains repeated sources. Index the $B_v$ sampled entries by $b$, with
source and modification indices $(i_b,k_b)$, where $k_b\geq1$. Let
\[
\mathcal J_i=\{b:i_b=i\},\qquad
\mathcal G=\{i:|\mathcal J_i|\geq2\}.
\]
For each eligible source, the pooled query is
\begin{equation}
z_i=\operatorname{normalize}\left(
\frac{1}{|\mathcal J_i|}\sum_{b\in\mathcal J_i}q_{i_b,k_b}\right).
\end{equation}
The cardinality $|\mathcal J_i|$ counts sampled occurrences, which may
include repeated draws of a modification. Identity queries are excluded.
The teacher bank retains the source-image embedding for every sampled
entry, including repeated occurrences. With unit-normalized teacher
embeddings, the positive positions for source $i$ are
\begin{equation}\label{eq:visual_positives}
\mathcal P_i=\{b:i_b=i\ \text{or}\ r_i^\top r_{i_b}>0.999\}.
\end{equation}
The implemented objective is
\begin{equation}\label{eq:visual_full}
\mathcal L_{\rm visual}
=-\frac{1}{|\mathcal G|}\sum_{i\in\mathcal G}
\log\frac{\sum_{b\in\mathcal P_i}\exp(z_i^\top r_{i_b}/\tau_v)}
{\sum_{b=1}^{B_v}\exp(z_i^\top r_{i_b}/\tau_v)}.
\end{equation}
Thus, repeated or near-identical source images contribute positive mass
rather than being treated as negatives. Sources represented once remain
in the teacher bank but do not contribute pooled queries. The visual loss
is zero when $\mathcal G$ is empty.

\section{Evaluation protocols}\label{app:evaluation}\label{app:protocol}

\subsection{Benchmarks and metrics}
CIRR \citep{liu2021cirr} and CIRCO \citep{baldrati2023zero} use official
test evaluation, while FashionIQ \citep{wu2019fashioniq} and GeneCIS
\citep{vaze2023genecis} use validation evaluation. CIRR excludes the
reference image and uses global top-50 and subset top-3 predictions.
CIRCO uses its 123,403-image gallery and top-50 predictions. In our local
CIRCO evaluation, the reference image remains in the candidate gallery.
FashionIQ retrieval is restricted to the query category. We average query
representations from the two caption orders before ranking, and report
the unweighted mean of Dress, Toptee, and Shirt recall.

GeneCIS evaluates conditional similarity under instructions that focus on or
change objects and attributes. Table~\ref{tab:requested3} reports recall
at ranks 1, 2, and 3 for the Focus, Change, Attribute, and Object summaries.

\subsection{Sources of GeneCIS baseline results}\label{app:genecis_sources}
The GeneCIS comparison combines local evaluation of available checkpoints
with published results, as summarized in Table~\ref{tab:genecis_sources}.
We evaluate Pic2Word, SEARLE-XL, and LinCIR using their released
checkpoints. For RTD, we evaluate a checkpoint reproduced using the
official code. CIReVL summaries are computed from the task-wise scores
reported in its paper. MoTa-Adapter provides an R@1 summary of 16.7,
without the task marginals shown for the other methods. DiffComp uses
the GeneCIS results reported in its paper. Missing metrics remain
unreported; published rows are not described as local reproductions.

\begin{table}[htbp]
\centering\small
\caption{\textbf{Sources of GeneCIS baseline results.}}
\label{tab:genecis_sources}
\begin{tabularx}{\linewidth}{lX}
\toprule
Method & Evaluation source\\
\midrule
Pic2Word \citep{saito2023pic2word} & Local evaluation of the authors' released checkpoint\\
SEARLE-XL \citep{baldrati2023zero} & Local evaluation of the authors' released checkpoint\\
LinCIR \citep{gu2024lincir} & Local evaluation of a released checkpoint\\
RTD \citep{byun2025rtd} & Local evaluation of a checkpoint reproduced with official code\\
CIReVL \citep{karthik2024cirevl} & Summaries computed from published task-wise scores\\
MoTa-Adapter \citep{li2026mota} & Published R@1 summary (16.7)\\
DiffComp \citep{zhang2026diffcomp} & Published GeneCIS results\\
\bottomrule
\end{tabularx}
\end{table}

\subsection{Checkpoint and scoring-weight selection}
We select checkpoints and the Chord coefficient using the unweighted
validation mean of CIRR R@1, CIRCO mAP@5, and FashionIQ category-average
R@10. The coefficient is selected from
$\lambda\in\{0,0.02,\ldots,1\}$. For each PACT training seed, we select
one coefficient that maximizes this validation mean and use it across
datasets; test scores do not enter selection. GeneCIS is excluded from
selection and uses the same coefficient without further tuning.

The benchmark tables compare complete retrieval systems, with GeneCIS
baseline sources specified in Table~\ref{tab:genecis_sources}. The Chord analysis instead
compares scoring rules while holding each query model and its gallery
embeddings fixed. For the training ablations, each configuration selects
its coefficient on the same validation grid. The cosine and Chord rows
in Table~\ref{tab:ablation} use matched checkpoints, separating the effect
of scoring from changes in the trained model.

\subsection{Variability across training seeds}
Table~\ref{tab:seed_std} reports variation across five independent training
runs for each backbone. All entries are validation results with Chord
scoring. We report sample standard deviations across five seeds,
in percentage points.
For FashionIQ, we first average the three category scores within each run
and then compute the standard deviation across runs. The GeneCIS entry
similarly uses the four-task mean for each run. These standard deviations
measure training-seed variation and are distinct from the query-sampling
bootstrap intervals in Appendix~\ref{app:chordanalysis}.

\begin{table}[htbp]
\centering\small
\caption{\textbf{Standard deviation across five training seeds.}
Validation results with Chord scoring. Values are sample standard
deviations, in percentage points.}
\label{tab:seed_std}
\begin{tabular}{lrrr}
\toprule
Benchmark and metric & ViT-B/32 & ViT-L/14 & ViT-G/14\\
\midrule
CIRR R@1 & 0.268 & 0.386 & 0.319\\
CIRCO mAP@5 & 0.344 & 1.072 & 0.592\\
FashionIQ category-average R@10 & 0.115 & 0.340 & 0.327\\
FashionIQ category-average R@50 & 0.298 & 0.476 & 0.395\\
GeneCIS four-task mean R@1 & 0.144 & 0.384 & 0.434\\
\bottomrule
\end{tabular}
\end{table}

Across the metrics and backbones in Table~\ref{tab:seed_std}, standard deviations range from
0.115 to 1.072 percentage points. CIRCO mAP@5 has the largest standard
deviation for each backbone, while the FashionIQ category-average R@10
standard deviation ranges from 0.115 to 0.340 points.

\clearpage
\subsection{Additional backbone results}
Tables~\ref{tab:backbone_compare_b} and~\ref{tab:backbone_compare_g}
report PACT performance with ViT-B/32 and ViT-G/14 alongside published
baselines; Table~\ref{tab:backbone_five_seed} provides additional retrieval
metrics. PACT uses Chord scoring. We use seeds 1007, 1013, 1019, 1021,
and 1031 and report mean $\pm$ sample standard deviation
across five seeds. CIRR and CIRCO results are
evaluated on their test sets, and FashionIQ results on its validation set.
For FashionIQ, each seed's score is the equally weighted mean across
the three clothing categories; the reported mean and standard deviation
are then computed across these five seed-level scores. GeneCIS summaries
likewise average the four tasks within each seed before aggregation
across seeds. The Chord coefficient is selected separately for each seed
using only the three-benchmark validation objective described above.

\begin{table}[H]
\centering\small
\caption{\textbf{Additional PACT retrieval metrics across five training seeds.}
Chord scoring; means are in percent and sample standard deviations are
in percentage points.}
\label{tab:backbone_five_seed}
\begin{tabular}{llcc}
\toprule
Benchmark & Metric & ViT-B/32 & ViT-G/14\\
\midrule
CIRR test & R@50 & $92.34\pm0.19$ & $95.68\pm0.05$\\
& Subset R@1 & $63.46\pm0.65$ & $70.46\pm0.43$\\
& Subset R@2 & $82.42\pm0.38$ & $87.22\pm0.35$\\
& Subset R@3 & $91.24\pm0.16$ & $93.61\pm0.17$\\
\midrule
CIRCO test & mAP@50 & $19.94\pm0.23$ & $35.70\pm0.75$\\
\bottomrule
\end{tabular}
\end{table}

\subsection{Category-wise FashionIQ results}
Table~\ref{tab:fiq_five_seed} reports category-wise FashionIQ validation
recall with ViT-L/14 and Chord scoring. Each entry is the mean and sample standard
deviation across five training seeds. The category-average row summarizes
the equally weighted category mean computed within each run.

\begin{table}[H]
\centering\small
\caption{\textbf{FashionIQ validation across five training seeds.}
ViT-L/14 with Chord scoring; mean $\pm$ sample standard deviation, in percentage points.}
\label{tab:fiq_five_seed}
\begin{tabular}{lcc}
\toprule
Category & R@10 & R@50\\
\midrule
Dress & $27.86\pm0.49$ & $49.70\pm0.42$\\
Shirt & $34.15\pm0.53$ & $52.25\pm0.20$\\
Toptee & $35.28\pm0.30$ & $54.94\pm0.89$\\
\midrule
Category average & $32.43\pm0.34$ & $52.29\pm0.48$\\
\bottomrule
\end{tabular}
\end{table}

\clearpage
\subsection{Comparison with published results across backbones}
Tables~\ref{tab:backbone_compare_b} and~\ref{tab:backbone_compare_g}
extend the comparison in Table~\ref{tab:requested1} to ViT-B/32 and
ViT-G/14 with the same metrics. Baselines retain the training and scoring
procedures of their published evaluations. PACT uses Chord scoring and
reports means and sample standard deviations across five seeds; additional
metrics are given in Table~\ref{tab:backbone_five_seed}. We retain all baseline
methods from Table~\ref{tab:requested1}; -- indicates that no published
value or local reproduction is available for that backbone and metric.
SEARLE denotes the corresponding backbone variant of SEARLE-XL.
RTD and MoTa-Adapter in Table~\ref{tab:requested1} correspond to the
LinCIR-based variants explicitly named here.

\begin{table}[H]
\centering\scriptsize
\setlength{\tabcolsep}{3pt}
\caption{\textbf{CIRR/CIRCO test and FashionIQ validation, ViT-B/32.}
PACT cells show the five-seed mean above $\pm$ sample standard deviation
(\%p). Bold marks the highest mean in each column;
-- indicates no available result for this backbone and metric.}
\label{tab:backbone_compare_b}
\resizebox{\linewidth}{!}{%
\begin{tabular}{lrrrrrrrr}
\toprule
Method & \multicolumn{3}{c}{CIRR test} & \multicolumn{3}{c}{CIRCO test}
& \multicolumn{2}{c}{FashionIQ val}\\
\cmidrule(lr){2-4}\cmidrule(lr){5-7}\cmidrule(lr){8-9}
& R@1 & R@5 & R@10 & mAP@5 & mAP@10 & mAP@25 & Avg R@10 & Avg R@50\\
\midrule
Pic2Word \hyperlink{cite.saito2023pic2word}{(2023)}$^{\dagger}$ & 13.64 & 37.45 & 52.22 & 2.85 & 3.24 & 3.89 & 11.73 & 26.30\\
SEARLE \hyperlink{cite.baldrati2023zero}{(2023)}$^{\dagger}$ & 23.71 & 53.30 & 66.84 & 8.90 & 9.42 & 10.64 & 22.64 & 41.85\\
Context-I2W \hyperlink{cite.tang2024context}{(2024)} & -- & -- & -- & -- & -- & -- & -- & --\\
CIReVL \hyperlink{cite.karthik2024cirevl}{(2024)} & 23.94 & 52.51 & 66.00 & 14.94 & 15.42 & 17.00 & \textbf{28.29} & \textbf{49.35}\\
LinCIR \hyperlink{cite.gu2024lincir}{(2024)}$^{\dagger}$ & 18.87 & 45.66 & 58.43 & 6.25 & 6.74 & 7.62 & 18.14 & 36.20\\
PrediCIR \hyperlink{cite.tang2025predicir}{(2025)} & -- & -- & -- & -- & -- & -- & -- & --\\
MoA \hyperlink{cite.li2025moa}{(2025)} & -- & -- & -- & -- & -- & -- & -- & --\\
HIT \hyperlink{cite.li2025hit}{(2025)} & -- & -- & -- & -- & -- & -- & -- & --\\
Pic2Word + RTD \hyperlink{cite.byun2025rtd}{(2025)}$^{\dagger}$ & 23.59 & 51.76 & 65.16 & 6.39 & 6.66 & 7.64 & 22.50 & 42.86\\
SEARLE + RTD \hyperlink{cite.byun2025rtd}{(2025)}$^{\dagger}$ & 26.29 & 56.41 & 69.74 & 11.26 & 12.11 & 13.63 & 24.70 & 45.40\\
LinCIR + RTD \hyperlink{cite.byun2025rtd}{(2025)}$^{\dagger}$ & 24.82 & 53.47 & 66.87 & 8.94 & 9.35 & 10.57 & 22.79 & 43.68\\
LinCIR + MoTa-Adapter \hyperlink{cite.li2026mota}{(2026)} & -- & -- & -- & -- & -- & -- & -- & --\\
De-MINDS \hyperlink{cite.tang2026deminds}{(2026)} & -- & -- & -- & -- & -- & -- & -- & --\\
FlowCIR \hyperlink{cite.he2026flowcir}{(2026)} & 25.50 & 56.50 & 69.80 & 13.10 & 13.40 & 14.60 & 23.40 & 43.20\\
\midrule
\rowcolor{tint}
PACT (ours) & \shortstack[r]{\textbf{30.09}\\$\pm0.27$} & \shortstack[r]{\textbf{61.27}\\$\pm0.29$} & \shortstack[r]{\textbf{74.33}\\$\pm0.17$} & \shortstack[r]{\textbf{16.74}\\$\pm0.34$} & \shortstack[r]{\textbf{17.42}\\$\pm0.15$} & \shortstack[r]{\textbf{19.16}\\$\pm0.25$} & \shortstack[r]{27.15\\$\pm0.11$} & \shortstack[r]{47.05\\$\pm0.30$}\\
\bottomrule
\end{tabular}}
\vspace{3pt}
\begin{minipage}{\linewidth}\scriptsize $^{\dagger}$Results reported in RTD, supplementary Tables S6--S7 \citep{byun2025rtd}, including its evaluations of the base methods. CIReVL and FlowCIR results are from their respective Tables 1--2.\end{minipage}
\end{table}
\begin{table}[H]
\centering\scriptsize
\setlength{\tabcolsep}{3pt}
\caption{\textbf{CIRR/CIRCO test and FashionIQ validation, ViT-G/14.}
PACT cells show the five-seed mean above $\pm$ sample standard deviation
(\%p). Bold marks the highest mean in each column;
-- indicates no available result for this backbone and metric.}
\label{tab:backbone_compare_g}
\resizebox{\linewidth}{!}{%
\begin{tabular}{lrrrrrrrr}
\toprule
Method & \multicolumn{3}{c}{CIRR test} & \multicolumn{3}{c}{CIRCO test}
& \multicolumn{2}{c}{FashionIQ val}\\
\cmidrule(lr){2-4}\cmidrule(lr){5-7}\cmidrule(lr){8-9}
& R@1 & R@5 & R@10 & mAP@5 & mAP@10 & mAP@25 & Avg R@10 & Avg R@50\\
\midrule
Pic2Word \hyperlink{cite.saito2023pic2word}{(2023)}$^{\ddagger}$ & 30.41 & 58.12 & 69.23 & 5.54 & 5.59 & 6.68 & 31.28 & 51.89\\
SEARLE \hyperlink{cite.baldrati2023zero}{(2023)}$^{\ddagger}$ & 34.80 & 64.07 & 75.11 & 13.20 & 13.85 & 15.32 & 34.81 & 55.71\\
Context-I2W \hyperlink{cite.tang2024context}{(2024)} & -- & -- & -- & -- & -- & -- & -- & --\\
CIReVL \hyperlink{cite.karthik2024cirevl}{(2024)} & 34.65 & 64.29 & 75.06 & 26.77 & 27.59 & 29.96 & 32.19 & 52.36\\
LinCIR \hyperlink{cite.gu2024lincir}{(2024)}$^{\ddagger}$ & 35.25 & 64.72 & 76.05 & 19.71 & 21.01 & 23.13 & 45.11 & 65.69\\
PrediCIR \hyperlink{cite.tang2025predicir}{(2025)} & 37.00 & 66.10 & 77.90 & 23.70 & 24.60 & 25.40 & 47.20 & 67.80\\
MoA \hyperlink{cite.li2025moa}{(2025)} & -- & -- & -- & -- & -- & -- & -- & --\\
HIT \hyperlink{cite.li2025hit}{(2025)} & -- & -- & -- & -- & -- & -- & -- & --\\
Pic2Word + RTD \hyperlink{cite.byun2025rtd}{(2025)} & -- & -- & -- & -- & -- & -- & -- & --\\
SEARLE + RTD \hyperlink{cite.byun2025rtd}{(2025)} & -- & -- & -- & -- & -- & -- & -- & --\\
LinCIR + RTD \hyperlink{cite.byun2025rtd}{(2025)} & 36.31 & 67.47 & 78.31 & 21.08 & 22.29 & 24.46 & 46.21 & 67.26\\
LinCIR + MoTa-Adapter \hyperlink{cite.li2026mota}{(2026)} & 38.39 & 69.47 & 80.05 & 25.82 & 27.06 & -- & 47.06 & 67.37\\
De-MINDS \hyperlink{cite.tang2026deminds}{(2026)} & 40.30 & 69.60 & 79.80 & 30.80 & 32.20 & 33.90 & \textbf{49.50} & \textbf{70.50}\\
FlowCIR \hyperlink{cite.he2026flowcir}{(2026)} & -- & -- & -- & -- & -- & -- & -- & --\\
\midrule
\rowcolor{tint}
PACT (ours) & \shortstack[r]{\textbf{40.91}\\$\pm0.32$} & \shortstack[r]{\textbf{71.63}\\$\pm0.28$} & \shortstack[r]{\textbf{82.68}\\$\pm0.21$} & \shortstack[r]{\textbf{31.48}\\$\pm0.59$} & \shortstack[r]{\textbf{32.78}\\$\pm0.73$} & \shortstack[r]{\textbf{34.92}\\$\pm0.75$} & \shortstack[r]{46.43\\$\pm0.33$} & \shortstack[r]{66.45\\$\pm0.39$}\\
\bottomrule
\end{tabular}}
\vspace{3pt}
\begin{minipage}{\linewidth}\scriptsize $^{\ddagger}$Results reported in LinCIR, Tables 3, 5, and 6 \citep{gu2024lincir}. LinCIR + RTD uses RTD supplementary Tables S8--S9; LinCIR + MoTa-Adapter uses its Table 1. CIReVL, PrediCIR, and De-MINDS use their original reported results.\end{minipage}
\end{table}

Among the compared methods, PACT achieves the highest CIRR R@1/5/10
and CIRCO mAP@5/10/25 with both backbones. FashionIQ shows a different
pattern: CIReVL leads with ViT-B/32, while De-MINDS leads with ViT-G/14.

\subsection{GeneCIS comparison across backbones}
We report the available GeneCIS results at their published level of
aggregation. We retain the baseline methods from Table~\ref{tab:requested3}
and leave unavailable metrics as --. Table~\ref{tab:genecis_b_average}
uses the four-task averages
reported in RTD, supplementary Table S4 \citep{byun2025rtd}; task-level
scores are not reported there. These six baseline rows retain RTD's evaluation
setting and are not combined with SEARLE results from other evaluations.
PACT's average assigns equal weight to Focus Attribute, Change Attribute,
Focus Object, and Change Object and is computed across the five seeds.

\begin{table}[H]
\centering\small
\caption{\textbf{GeneCIS four-task averages with ViT-B/32.}
RTD-evaluated baselines and a separately sourced CIReVL row. PACT uses
Chord scoring and five-seed means. Bold marks column maxima;
-- indicates no available result for this backbone and metric.}
\label{tab:genecis_b_average}
\begin{tabular}{lrrr}
\toprule
Method & Average R@1 & Average R@2 & Average R@3\\
\midrule
Pic2Word \hyperlink{cite.saito2023pic2word}{(2023)} & 11.13 & 21.08 & 31.05\\
SEARLE \hyperlink{cite.baldrati2023zero}{(2023)} & 12.19 & 22.56 & 32.03\\
CIReVL \hyperlink{cite.karthik2024cirevl}{(2024)}$^{\ddagger}$ & 15.85 & 26.83 & 36.78\\
LinCIR \hyperlink{cite.gu2024lincir}{(2024)} & 12.23 & 21.29 & 30.80\\
Pic2Word + RTD \hyperlink{cite.byun2025rtd}{(2025)} & 12.03 & 21.61 & 31.09\\
SEARLE + RTD \hyperlink{cite.byun2025rtd}{(2025)} & 12.82 & 22.97 & 32.44\\
LinCIR + RTD \hyperlink{cite.byun2025rtd}{(2025)} & 12.83 & 22.83 & 32.22\\
MoTa-Adapter \hyperlink{cite.li2026mota}{(2026)} & -- & -- & --\\
DiffComp \hyperlink{cite.zhang2026diffcomp}{(2026)} & -- & -- & --\\
\midrule
\rowcolor{tint}
PACT (ours) & \textbf{17.23} & \textbf{28.53} & \textbf{38.65}\\
\bottomrule
\end{tabular}
\vspace{3pt}
\begin{minipage}{\linewidth}\scriptsize
$^{\ddagger}$Equal averages of the four ViT-B/32 task scores in CIReVL,
Table 3 \citep{karthik2024cirevl}; this row uses CIReVL's evaluation
setting. Pic2Word, SEARLE, LinCIR, and their +RTD rows come from RTD Table S4.
\end{minipage}
\end{table}

Table~\ref{tab:genecis_g_tasks} reports the four individual tasks for
ViT-G/14. Pic2Word, SEARLE, and LinCIR results are reported in LinCIR,
supplementary Table B.3 \citep{gu2024lincir}; CIReVL results are reported
in its Table 3 \citep{karthik2024cirevl}. The task order follows Table B.3:
Focus Attribute, Change Attribute, Focus Object, and Change Object.
These are individual tasks, rather than the overlapping marginals in
Table~\ref{tab:requested3}. PACT entries are five-seed means with Chord
scoring. PACT has the highest mean in each displayed task and recall rank
among these comparison methods.

\begin{table}[H]
\centering\scriptsize\setlength{\tabcolsep}{3pt}
\caption{\textbf{GeneCIS task-level results with ViT-G/14.}
Published baselines and five-seed PACT means. Bold indicates the highest
value in each column among the compared methods; -- indicates no
available task-level result.}
\label{tab:genecis_g_tasks}
\resizebox{\linewidth}{!}{%
\begin{tabular}{lrrrrrrrrrrrr}
\toprule
Method & \multicolumn{3}{c}{Focus Attribute} & \multicolumn{3}{c}{Change Attribute}
& \multicolumn{3}{c}{Focus Object} & \multicolumn{3}{c}{Change Object}\\
\cmidrule(lr){2-4}\cmidrule(lr){5-7}\cmidrule(lr){8-10}\cmidrule(lr){11-13}
& R@1 & R@2 & R@3 & R@1 & R@2 & R@3 & R@1 & R@2 & R@3 & R@1 & R@2 & R@3\\
\midrule
Pic2Word \hyperlink{cite.saito2023pic2word}{(2023)} & 12.45 & 23.40 & 33.65 & 11.74 & 21.88 & 30.87 & 9.90 & 19.34 & 27.35 & 8.57 & 18.16 & 26.12\\
SEARLE \hyperlink{cite.baldrati2023zero}{(2023)} & 16.30 & 29.40 & 40.70 & 16.15 & 27.32 & 35.46 & 10.77 & 18.16 & 27.91 & 8.27 & 15.56 & 25.77\\
CIReVL \hyperlink{cite.karthik2024cirevl}{(2024)} & 20.50 & 34.00 & 44.50 & 16.10 & 28.60 & 39.40 & 14.70 & 25.20 & 33.00 & 18.10 & 31.20 & 41.00\\
LinCIR \hyperlink{cite.gu2024lincir}{(2024)} & 19.05 & 33.00 & 42.30 & 17.57 & 30.16 & 38.07 & 10.10 & 19.08 & 28.06 & 7.91 & 16.33 & 25.71\\
RTD \hyperlink{cite.byun2025rtd}{(2025)} & -- & -- & -- & -- & -- & -- & -- & -- & -- & -- & -- & --\\
MoTa-Adapter \hyperlink{cite.li2026mota}{(2026)} & -- & -- & -- & -- & -- & -- & -- & -- & -- & -- & -- & --\\
DiffComp \hyperlink{cite.zhang2026diffcomp}{(2026)} & -- & -- & -- & -- & -- & -- & -- & -- & -- & -- & -- & --\\
\midrule
\rowcolor{tint}
PACT (ours) & \textbf{22.32} & \textbf{34.30} & \textbf{45.65} & \textbf{19.11} & \textbf{32.68} & \textbf{42.11} & \textbf{20.22} & \textbf{29.12} & \textbf{37.77} & \textbf{19.83} & \textbf{32.72} & \textbf{42.06}\\
\bottomrule
\end{tabular}}
\vspace{3pt}
\begin{minipage}{\linewidth}\scriptsize
LinCIR + MoTa-Adapter reports a four-task R@1 summary of 19.1
\citep{li2026mota}, but no task-level values; its task cells remain --.
\end{minipage}
\end{table}

\section{Additional analysis of Chord scoring}\label{app:chordanalysis}

\subsection{Retrieval implementation and per-dataset gains}
Chord operates on unit-normalized query and gallery representations.
Each model constructs its identity query with an empty modification.
The reference-relative term is set to zero for a candidate identical to
the reference image. Gallery embeddings remain fixed, and scoring uses
no inverted-softmax calibration. The reference is excluded on CIRR,
retained in local CIRCO evaluation, and FashionIQ candidates are restricted
to the relevant category.

Table~\ref{tab:chord_detail} expands the mean gains in
Figure~\ref{fig:intrinsic}a into individual benchmark results.
Pic2Word and LinCIR use released mappers; RTD is reproduced with its
official implementation. Within each model, cosine and Chord use the same
query representations and candidate embeddings.

\begin{table}[htbp]
\centering\small\setlength{\tabcolsep}{5pt}
\caption{\textbf{Cosine and Chord retrieval on validation data.}
Entries show cosine $\rightarrow$ Chord with one selected coefficient per
model. PACT corresponds to the full-model rows in Table~\ref{tab:ablation}.}
\label{tab:chord_detail}
\begin{tabular}{lrrrr}
\toprule
Model & $\lambda$ & CIRR R@1 & CIRCO mAP@5 & FashionIQ R@10\\
\midrule
Pic2Word & 0.16 & $22.82\to23.44$ & $8.10\to9.02$ & $25.56\to25.76$\\
LinCIR & 0.34 & $25.09\to27.36$ & $10.65\to12.90$ & $26.05\to27.30$\\
RTD & 0.08 & $27.12\to27.77$ & $13.23\to14.03$ & $29.93\to29.83$\\
\rowcolor{tint}
PACT & 0.34 & $25.76\to32.50$ & $18.68\to24.18$ & $28.29\to32.43$\\
\bottomrule
\end{tabular}
\end{table}

PACT improves on all three benchmarks, by 6.74 points on CIRR, 5.50 on
CIRCO, and 4.14 on FashionIQ. Chord also increases the aggregate performance
of the other query models, although gains need not occur on every dataset:
RTD improves on CIRR and CIRCO and decreases by 0.10 points on FashionIQ.
These results reflect validation-selected scoring coefficients.

\subsection{Cosine ambiguity and pair correction}
For each query, we compare every annotated positive with the highest-cosine
candidate that is not annotated as positive. We retain pairs for which
\[
\delta=\cos(q,g_n)-\cos(q,g_p)>10^{-7}.
\]
Each query contributes equal total pair weight within its category, and
FashionIQ categories receive equal weight. Within each benchmark, we pool
the four models' cosine-deficit distributions with equal weight per model
and form five weighted quantile bins. The boundaries are shared across
models. Q1 contains the largest deficits and Q5 the smallest. These bins
measure cosine-ranking ambiguity. On CIRCO, candidates outside the
annotated positive set are unannotated, rather than verified irrelevant.

Within each quintile, we stratify the positive's cosine rank into
$[2,6)$, $[6,11)$, $[11,21)$, $[21,51)$, $[51,101)$, $[101,501)$, and
$[501,\infty)$. A rank cell is retained only if it contains at least ten
distinct queries for every model. We use common weights proportional to
the minimum query-weighted cell mass across models to standardize the rank
distribution. This controls the rank composition of the comparison;
the models can still contribute different queries and candidates.

\begin{figure}[H]
\centering
\includegraphics[width=\linewidth]{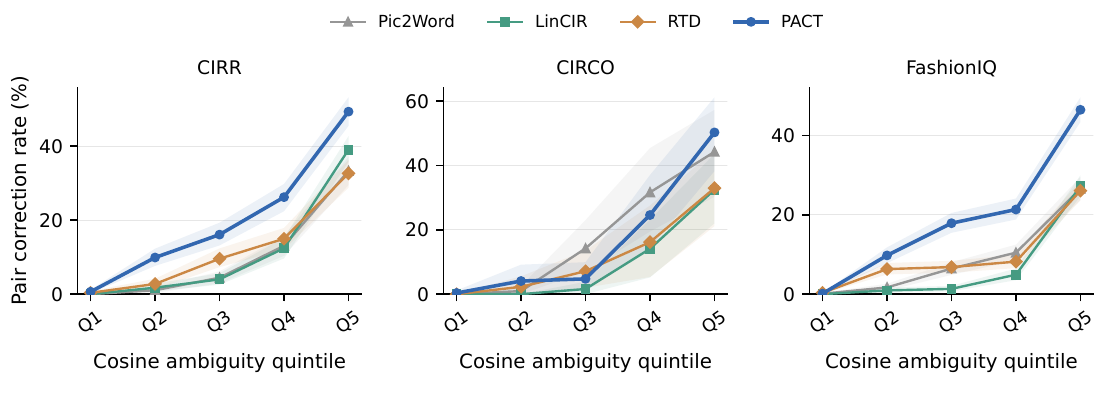}
\caption{\textbf{Chord pair correction by benchmark.} Shared cosine-deficit
quintiles and rank standardization follow the procedure described above.
All models use $\lambda=0.34$. Q5 contains the smallest deficits; shading
denotes 95\% reference-cluster bootstrap intervals.}
\label{fig:chord_by_dataset}
\end{figure}

We measure the fraction of misordered pairs corrected by Chord at a common
$\lambda=0.34$, and average the three benchmark rates equally for
Figure~\ref{fig:intrinsic}b. The 95\% intervals use 1,000 bootstrap
replicates clustered by reference image, with bin boundaries, support,
and standardization weights held fixed. They quantify query-sampling
variation rather than variation across training seeds.
Figure~\ref{fig:chord_by_dataset} reports the individual benchmark curves.

\section{Reference preservation on MovieNet}\label{app:movienet}

\subsection{Evaluation sets}
We use MovieNet \citep{huang2020movienet} for two complementary evaluations:
an annotation-based identity-preservation evaluation and qualitative
retrieval over a larger gallery. Table~\ref{tab:movienet_sets} summarizes
their respective query and gallery sets. This section describes the
quantitative evaluation in Table~\ref{tab:movienet_preserve}; the
qualitative comparison is described in Appendix~\ref{app:qualitative}.

\begin{table}[htbp]
\centering\small
\caption{\textbf{MovieNet evaluation sets.} Query-film counts refer to the
films represented by the references; gallery-film counts refer to the
retrieval collections.}
\label{tab:movienet_sets}
\begin{tabular}{lrr}
\toprule
 & Identity-preservation evaluation & Qualitative comparison\\
\midrule
Queries & 200 & 1,151\\
Distinct reference images & 200 & 90\\
Films represented in queries & 62 & 4\\
Gallery images & 6,162 & 100,000\\
Films represented in gallery & 66 & 104\\
\bottomrule
\end{tabular}
\end{table}

\subsection{Gallery and query construction}
From the 100,000-image MovieNet collection, we retain frames with official
person annotations for which every annotated person has a single,
unambiguous identity assignment. Frames with unresolved or multiple
candidate identities are excluded. The resulting gallery contains 6,162
frames from 66 films and includes images with multiple people.

To construct queries, we form source--target pairs from the same film
and with the same person identity. Each frame in a pair has one annotated
person, the files have different contents, and the shot indices differ
by at least 40. Qwen3.5-27B receives both images and generates an
instruction that preserves the person while changing the background,
scene, or action. Automatic verification checks that the target satisfies
the instruction and the source does not already satisfy it. Reference
images undergo additional checks for frontal or three-quarter face
orientation, clarity, illumination, and occlusion. The final set contains
200 queries with 200 distinct references from 62 films.

At retrieval, we exclude images with file contents identical to the
reference and images from the same shot in the same film. We retain
Top-5 predictions and evaluate the Top-1 result. Source--target pairs
are used to construct evaluation queries and do not provide training
supervision for PACT.

\subsection{Identity and modification criteria}
For query $i$, let $\hat G_i$ be the top-ranked image. Define $A_i=1$ when
the official MovieNet annotations for $\hat G_i$ contain the reference
person's film-ID/person-ID pair, and $A_i=0$ otherwise. The identity
preservation rate is
\begin{equation}\label{eq:movienet_identity}
\mathrm{Identity}=\frac{100}{N}\sum_{i=1}^{N} A_i,\qquad N=200.
\end{equation}
Identity is determined from official annotations, without face recognition
or language-model identity judgments. A retrieved image containing several
people is counted as preserving identity when the reference person's
annotation is present. This criterion measures person-identity retention;
it does not require preservation of every visual attribute, such as
clothing, color, or texture.

Modification fulfillment is assessed separately by Qwen3.5-27B, which
receives the reference image, retrieved image, and instruction with the
retrieval model's identity hidden. Let $M_i$ be its binary judgment of
whether the retrieved image fulfills the modification. We define
\begin{equation}
\mathrm{Modification}=\frac{100}{N}\sum_{i=1}^{N} M_i,\qquad
\mathrm{Joint}=\frac{100}{N}\sum_{i=1}^{N} A_iM_i.
\end{equation}
The joint criterion requires both identity retention and modification
fulfillment. Table~\ref{tab:movienet_preserve} reports the identity metric.

\subsection{Effect of visual supervision}
Adding the visual objective to composition-only training increases Top-1
identity from 58.0\% to 78.5\% with cosine and from 62.0\% to 81.5\% with
Chord, gains of 20.5 and 19.5 percentage points, respectively.
Full PACT reaches 80.5\% and 82.5\%, exceeding RTD by 16.5 and 18.0 points
under the corresponding scoring rules. The improvement under both rules
supports the role of visual supervision in retaining source-specific
identity.

\section{Qualitative retrieval on MovieNet}\label{app:qualitative}

\subsection{Gallery construction}
The qualitative evaluation uses 100,000 images from 104 films. The gallery
comprises 9,660 images from the four query films and 90,340 images from
100 additional films. The additional films are selected in a fixed hash
order from films with at least 1,000 shots, excluding the query films.
Shots are sampled evenly within each selected film, and the middle frame
of each sampled shot is used. Unlike the annotation-based collection in
Appendix~\ref{app:movienet}, this gallery is not restricted to frames with
unambiguous person-identity annotations.

\subsection{Reference images and modifications}
We use 90 reference images from \emph{The Matrix} (1999),
\emph{Pirates of the Caribbean: The Curse of the Black Pearl} (2003),
\emph{The Dark Knight} (2008), and \emph{Inception} (2010).
Descriptions of visual states are written after inspecting clothing,
pose, action, and background. Modifications request a different state of
the same person, using combinations of these conditions. Additional
instructions isolate subsets of the described conditions, such as an
action or a background. The resulting set contains 1,151 queries, with
multiple modifications associated with the same reference image.

\clearpage
\subsection{Baseline configurations}\label{app:qual_baselines}
The qualitative comparisons use the same 100,000-image MovieNet gallery
and ViT-L/14 backbone. Image-only and Instruction-only use the original,
frozen CLIP encoders; neither passes through the learned PACT mapper or
adapted text encoder. Compose-only is a training-objective ablation of
PACT and retains both query modalities. Let $v(I)$ and $t(m)$ denote
the $L_2$-normalized embeddings from the original CLIP image and text
encoders, respectively, and let $I_g$ be a gallery image.

\paragraph{Image-only.}
The reference image is encoded by the original CLIP image encoder, and
its normalized embedding is used directly as the retrieval query:
\begin{equation}
s_{\text{image-only}}(I_g)=v(I_{\mathrm{ref}})^\top v(I_g).
\end{equation}
This cosine baseline receives no modification instruction and uses
neither the PACT mapper nor its adapted text encoder.
It measures retrieval from source appearance alone.

\paragraph{Instruction-only.}
The complete, unmodified instruction $m$ is encoded by the original
frozen CLIP text encoder and normalized before cosine retrieval:
\begin{equation}
s_{\text{instruction-only}}(I_g)=t(m)^\top v(I_g).
\end{equation}
The input includes reference-dependent phrases such as ``Keep the same
character, but \ldots'' exactly as written. No reference image,
source-image caption, or image-dependent pseudo-word is supplied.
The learned PACT text encoder is not used.
Both single-modality baselines use the same gallery image embeddings
and exclude the reference image and its byte-identical duplicates
from retrieval.

\paragraph{Compose-only.}
This model uses the same training data, architecture, trainable modules,
and optimization schedule as full PACT (Appendix~\ref{app:implementation}).
It retains the 333-update identity warmup and 1,000-update joint-training
stage. The mapper and text-attention LoRA follow
the same training procedure; the image encoder and teacher text encoder
remain frozen. Only the joint objective changes: it is
$\mathcal L_{\rm compose}$ from Equation~\ref{eq:compose}, with the visual
and negative-caption preservation terms disabled. Thus, the model still
receives identity-warmup supervision and both query modalities.
At retrieval time, Compose-only uses Chord with $\lambda=0.34$, the same
scoring rule and coefficient as full PACT. Their comparison therefore
holds the query inputs and retrieval scoring fixed while varying the
joint-training objective.

\paragraph{Full PACT and RTD.}
Full PACT uses both the reference image and modification instruction,
the complete training objective, and Chord with $\lambda=0.34$.
RTD uses the official-code reproduction checkpoint and cosine scoring.
All displayed model results are the original Top-1 retrievals for the
corresponding query and checkpoint.

\subsection{Comparison and interpretation}
\begin{figure}[H]
\centering
\includegraphics[width=\linewidth]{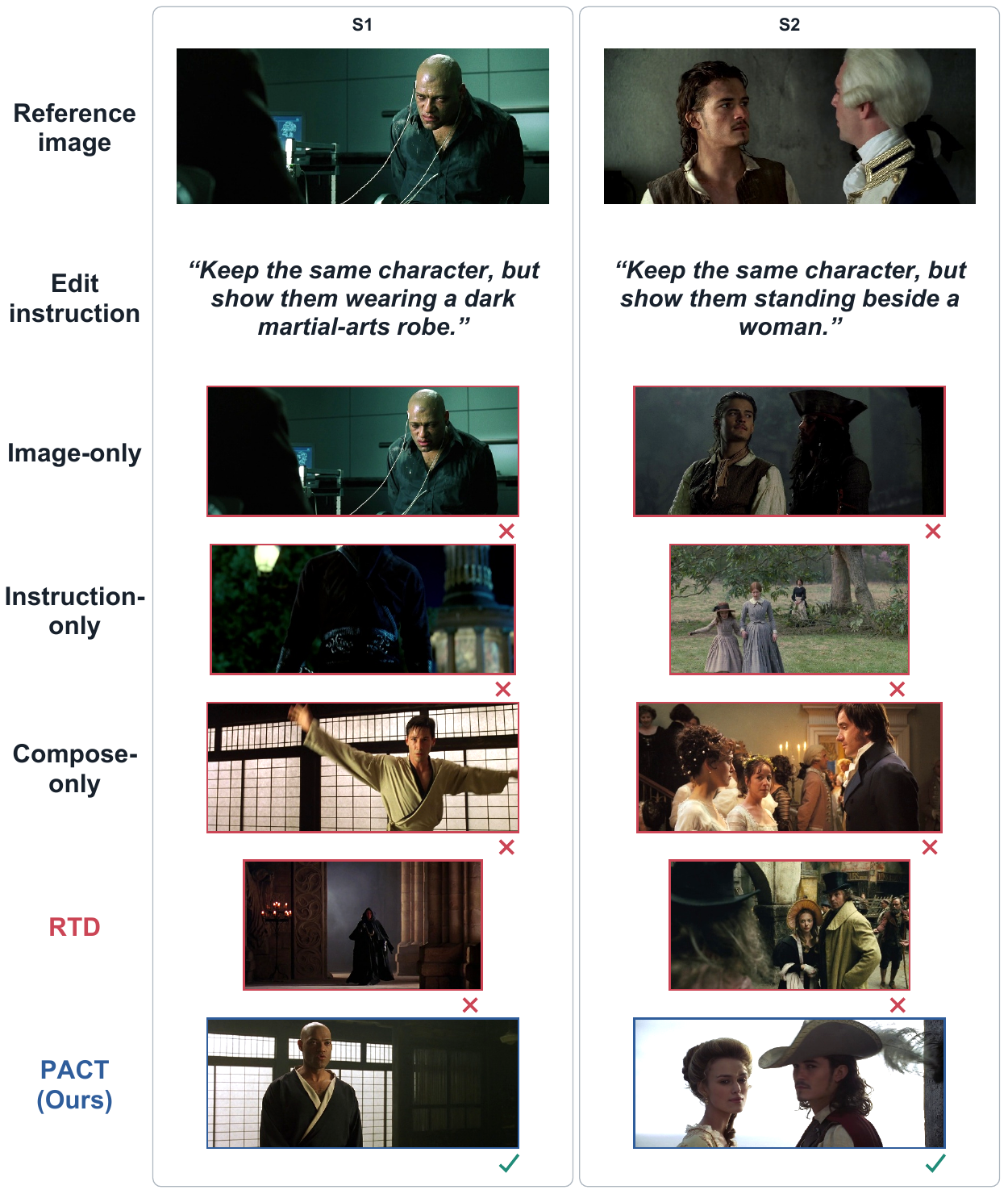}
\caption{MovieNet Top-1 comparisons S1--S2. S1: composition-only retrieves a different character in a pale outfit. S2: it matches the beside-a-woman relation but loses the reference character. PACT satisfies both requests. ``Compose-only'' denotes $\mathcal L_{\rm compose}$-only training.}
\label{fig:qual_ablation}
\end{figure}

\begin{figure}[H]
\centering
\includegraphics[width=\linewidth]{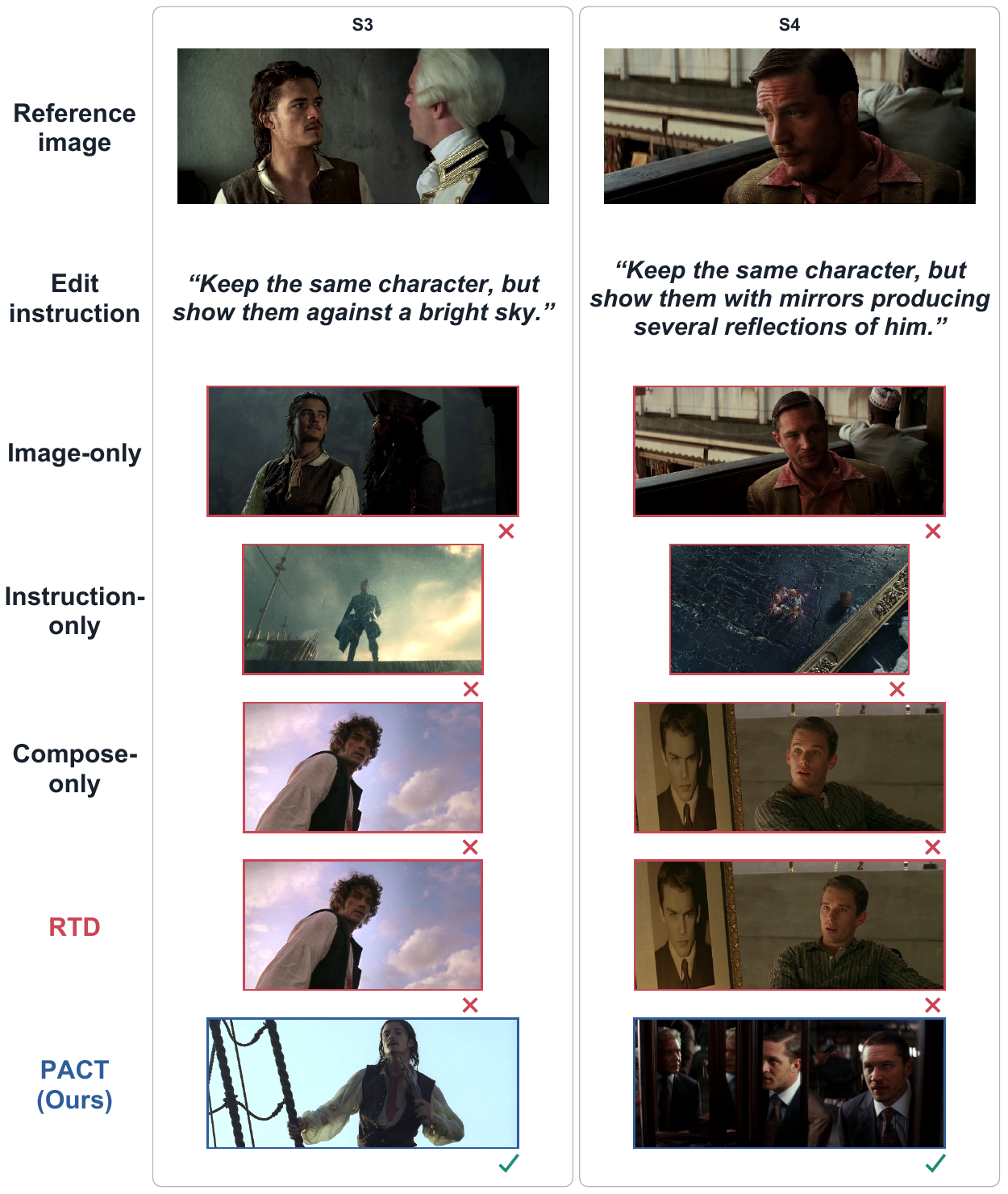}
\caption{MovieNet Top-1 comparisons S3--S4. S3: composition-only matches the bright sky with a different character. S4: it retrieves a portrait instead of mirror reflections. PACT satisfies both requests. ``Compose-only'' denotes $\mathcal L_{\rm compose}$-only training.}
\label{fig:qual_success_pair_2}
\end{figure}

\begin{figure}[H]
\centering
\includegraphics[width=\linewidth]{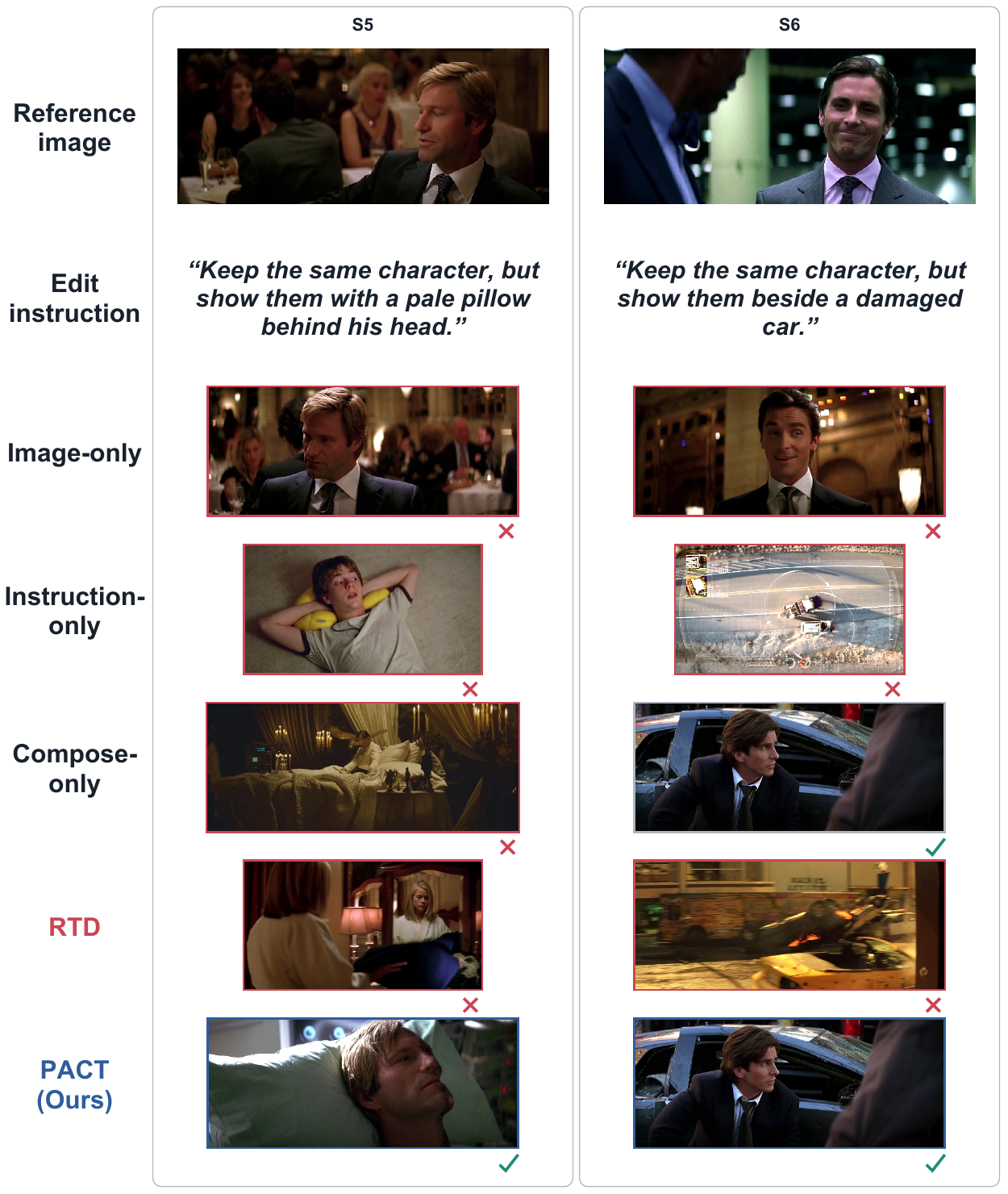}
\caption{MovieNet Top-1 comparisons S5--S6. S5: PACT retrieves the requested pale pillow behind the head. S6: both composition-only and PACT retrieve the reference character beside a damaged car. ``Compose-only'' denotes $\mathcal L_{\rm compose}$-only training.}
\label{fig:qual_success_pair_3}
\end{figure}

\begin{figure}[H]
\centering
\includegraphics[width=\linewidth]{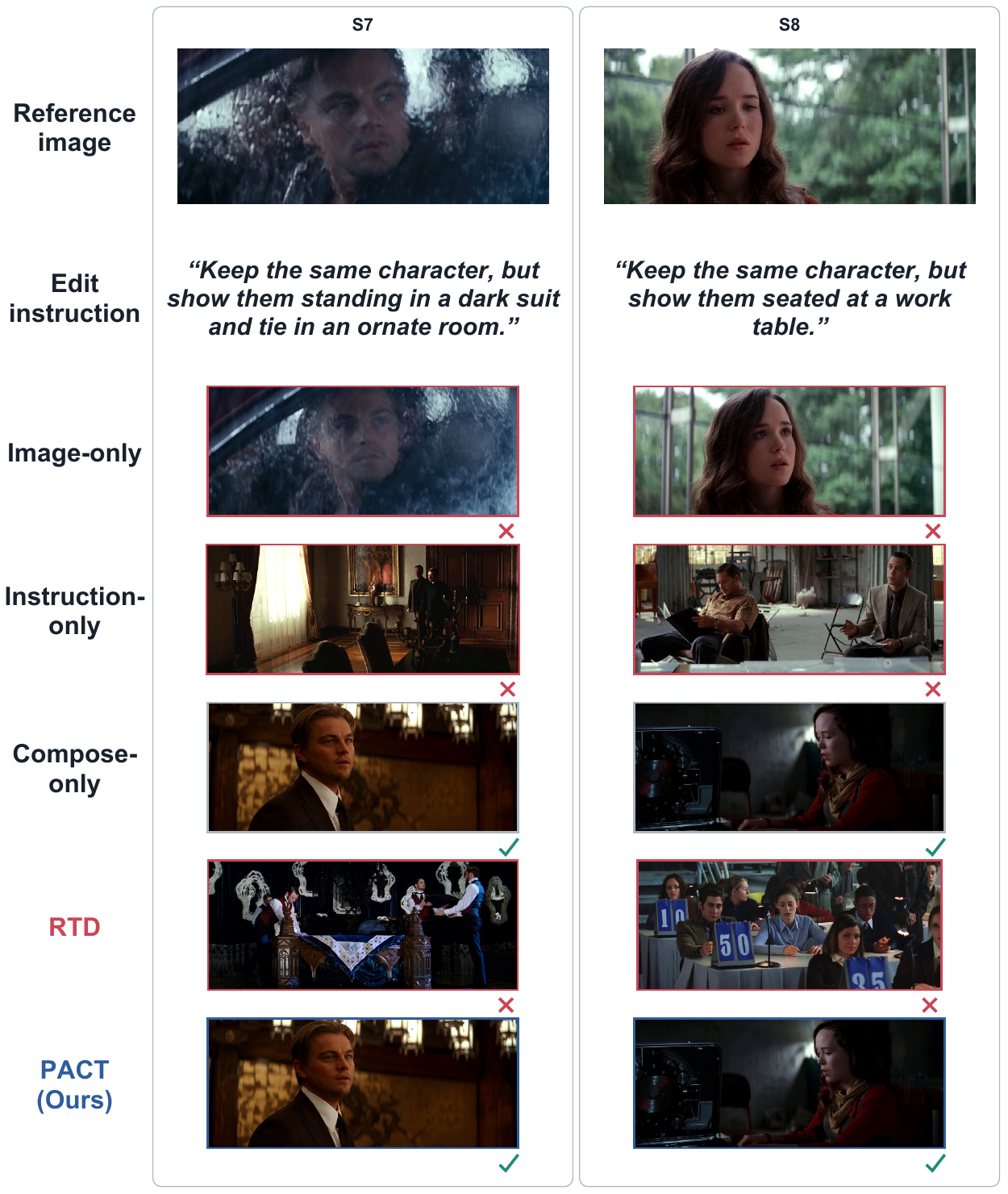}
\caption{MovieNet Top-1 comparisons S7--S8. Both composition-only and PACT satisfy the ornate-room (S7) and work-table (S8) requests. ``Compose-only'' denotes $\mathcal L_{\rm compose}$-only training.}
\label{fig:qual_success_last}
\end{figure}

Figures~\ref{fig:qual_ablation}--\ref{fig:qual_success_last} show eight
selected comparisons. Checks mark matched requests; crosses mark unmet
character or modification conditions. Blue identifies PACT independently
of correctness. A uniform 20\% brightness increase is applied only to
the displayed images for visibility, not to retrieval inputs. These
selected examples do not estimate overall success rates.

\clearpage

\clearpage

\clearpage

\clearpage

\clearpage
\subsection{Failure cases: fine-grained modification constraints}
\label{app:qual_failures}
\begin{figure}[H]
\centering
\includegraphics[width=\linewidth]{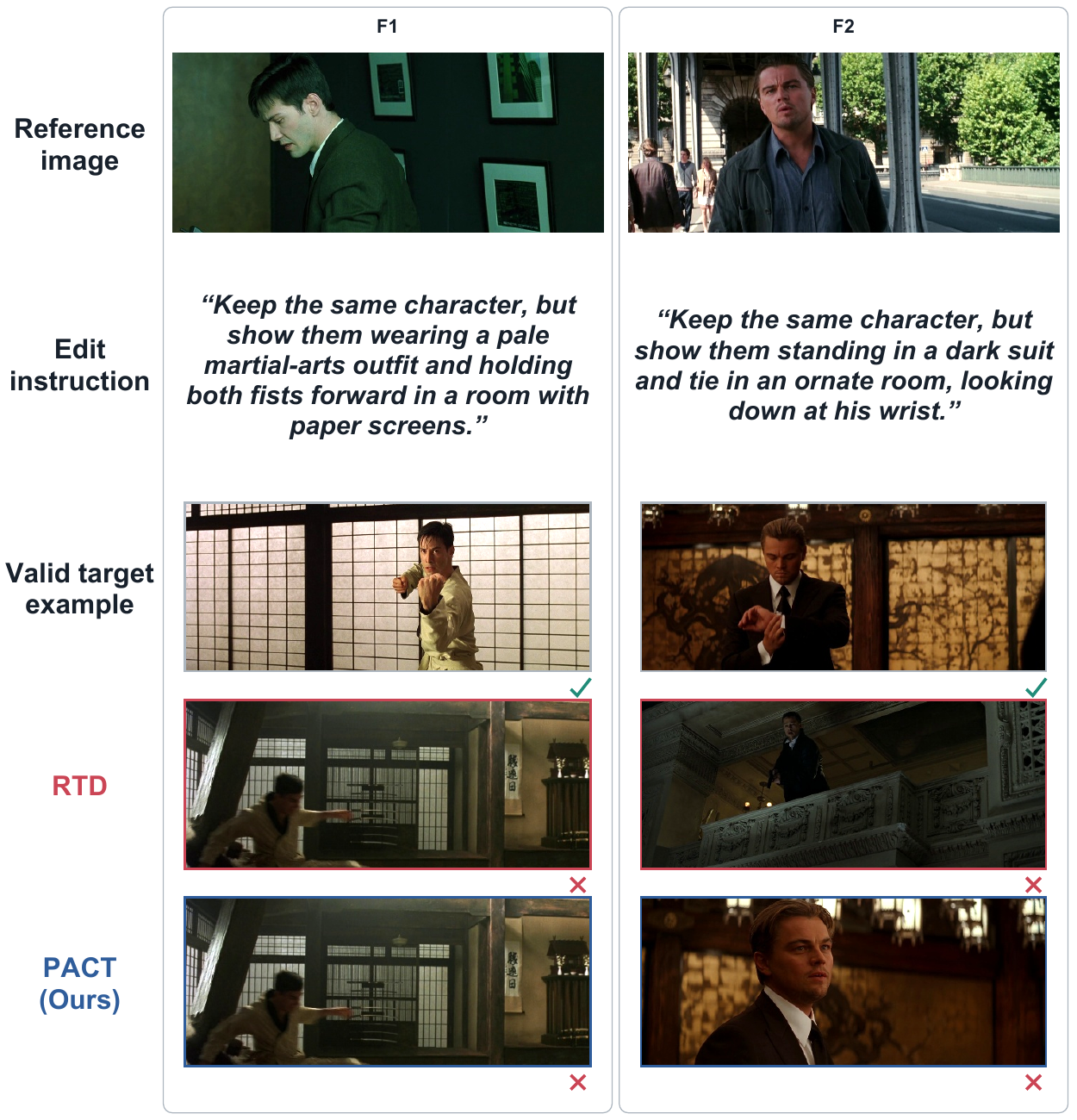}
\caption{MovieNet failures F1 and F2. In F1, both methods retrieve the same
martial-arts frame, but the requested two-fist pose is not visible. In F2,
both recover suit and interior cues without the action of looking down at
a wrist. Valid targets are illustrative gallery positives, not unique
correct frames. All model results are Top-1.}
\label{fig:qual_failures}
\end{figure}

Figure~\ref{fig:qual_failures} illustrates unmet pose and action
constraints. Both models use
ViT-L/14: PACT with Chord ($\lambda=0.34$), RTD with cosine. These
selected examples do not estimate failure frequency.

\clearpage
\section{Efficiency and model-size accounting}\label{app:parameter_counts}

\subsection{Training-time measurement}\label{app:trainingtime}
We measure the ViT-L/14 training configuration with batch size 512 on a
single RTX 3090. CUDA-synchronized updates for 333 identity-warmup and
1,000 joint-training steps take 52 min 9 s. Peak allocated GPU memory is
8.19 GiB. The timing excludes evaluation, checkpoint saving, offline data
generation, and image-feature preparation. These measurements characterize
optimization with cached supervision and frozen image features.

\subsection{Inference parameter accounting}
Parameter counts include the image and text encoders and query-side modules
used for retrieval; offline generation models and training-only teachers
are excluded. PACT-L/14 has approximately 443M inference parameters.
Figure~\ref{fig:params} uses CIRCO test mAP@5 as the common performance
measure. Architecture-derived estimates and component lower bounds are
distinguished below because they have different counting scopes.

For MoTa-Adapter \citep{li2026mota}, we add the adapter components to CLIP
and the LinCIR mapper. The added terms are $KNd_T+Kd_T+2d_I$, with $K=4$
and $N=8$, giving estimated totals of 441.81M for L/14 and 2.57895B for
G/14. For MCL \citep{li2024mcl}, the totals include the causal language
model, the CLIP-L/14 visual encoder, visual and retrieval projections,
and the fusion module. Tied weights are counted once. The OPT-2.7B,
OPT-6.7B, and Llama2-7B variants yield architecture-derived totals of
2.975B, 6.988B, and 7.068B, respectively.

CoLLM \citep{huynh2025collm} uses a nominal 7B language model; adding its
visual encoder gives a component lower bound because adapter and projection
overhead is not fully specified. Its plotted scores use image-caption
pretraining and exclude MTCIR fine-tuning. For PrediCIR
\citep{tang2025predicir}, the lower bound includes CLIP and the attention
projection weights of 12 predictor blocks of width 384:
$12(4\times384^2)=7{,}077{,}888$ additional parameters. Feed-forward layers
and other unspecified components are excluded, yielding lower bounds of
434.69M for L/14 and 2.54664B for G/14.

The parameter count of the GPT-3.5 component in CIReVL
\citep{karthik2024cirevl} is undisclosed. Its horizontal position in
Figure~\ref{fig:params} uses an illustrative 175B assumption for that
component, together with the captioner and CLIP. This position is not a
measured model size and cannot establish a quantitative parameter ratio
relative to PACT.

\end{document}